# Approximate Policy Iteration with a Policy Language Bias: Solving Relational Markov Decision Processes

**Alan Fern**                                                    AFERN@CS.ORST.EDU
*School of Electrical Engineering and Computer Science, Oregon State University*
**Sungwook Yoon**                                                      SY@PURDUE.EDU
**Robert Givan**                                                   GIVAN@PURDUE.EDU
*School of Electrical and Computer Engineering, Purdue University*

## Abstract

We study an approach to policy selection for large relational Markov Decision Processes (MDPs). We consider a variant of approximate policy iteration (API) that replaces the usual value-function learning step with a learning step in policy space. This is advantageous in domains where good policies are easier to represent and learn than the corresponding value functions, which is often the case for the relational MDPs we are interested in. In order to apply API to such problems, we introduce a relational policy language and corresponding learner. In addition, we introduce a new bootstrapping routine for goal-based planning domains, based on random walks. Such bootstrapping is necessary for many large relational MDPs, where reward is extremely sparse, as API is ineffective in such domains when initialized with an uninformed policy. Our experiments show that the resulting system is able to find good policies for a number of classical planning domains and their stochastic variants by solving them as extremely large relational MDPs. The experiments also point to some limitations of our approach, suggesting future work.

## 1. Introduction

Many planning domains are most naturally represented in terms of objects and relations among them. Accordingly, AI researchers have long studied algorithms for planning and learning-to-plan in relational state and action spaces. These include, for example, "classical" STRIPS domains such as the blocks world and logistics.

A common criticism of such domains and algorithms is the assumption of an idealized, deterministic world model. This, in part, has led AI researchers to study planning and learning within a decision-theoretic framework, which explicitly handles stochastic environments and generalized reward-based objectives. However, most of this work is based on explicit or propositional state-space models, and so far has not demonstrated scalability to the large relational domains that are commonly addressed in classical planning.

Intelligent agents must be able to simultaneously deal with both the complexity arising from relational structure and the complexity arising from uncertainty. The primary goal of this research is to move toward such agents by bridging the gap between classical and decision-theoretic techniques.

In this paper, we describe a straightforward and practical method for solving very large, relational MDPs. Our work can be viewed as a form of relational reinforcement learning (RRL) where we assume a strong simulation model of the environment. That is, we assume access to a black-box simulator, for which we can provide any (relationally represented) state/action pair and receive a sample from the appropriate next-state and reward distributions. The goal is to interact with the simulator in order to learn a policy for achieving high expected reward. It is a separate challenge, not considered here, to combine our work with methods for learning the environment simulator to avoid dependence on being provided such a simulator.

Dynamic-programming approaches to finding optimal control policies in MDPs (Bellman, 1957; Howard, 1960), using explicit (flat) state space representations, break down when the state space becomes extremely large. More recent work extends these algorithms to use propositional (Boutilier & Dearden, 1996; Dean & Givan, 1997; Dean, Givan, & Leach, 1997; Boutilier, Dearden, & Goldszmidt, 2000; Givan, Dean, & Greig, 2003; Guestrin, Koller, Parr, & Venkataraman, 2003b) as well as relational (Boutilier, Reiter, & Price, 2001; Guestrin, Koller, Gearhart, & Kanodia, 2003a) state-space representations. These extensions have significantly expanded the set of approachable problems, but have not yet shown the capacity to solve large classical planning problems such as the benchmark problems used in planning competitions (Bacchus, 2001), let alone their stochastic variants. One possible reason for this is that these methods are based on calculating and representing value functions. For familiar STRIPS planning domains (among others), useful value functions can be difficult to represent compactly, and their manipulation becomes a bottle-neck.

Most of the above techniques are purely deductive—that is, each value function is guaranteed to have a certain level of accuracy. Rather, in this work, we will focus on inductive techniques that make no such guarantees in practice. Most existing inductive forms of approximate policy iteration (API) utilize machine learning to select compactly represented approximate value functions at each iteration of dynamic programming (Bertsekas & Tsitsiklis, 1996). As with any machine learning algorithm, the selection of the hypothesis space, here a space of value functions, is critical to performance. An example space used frequently is the space of linear combinations of a human-selected feature set.

To our knowledge, there has been no previous work that applies any form of API to benchmark problems from classical planning, or their stochastic variants.[1] Again, one reason for this is the high complexity of typical value functions for these large relational domains, making it difficult to specify good value-function spaces that facilitate learning. Comparably, it is often much easier to compactly specify good policies, and accordingly good policy spaces for learning. This observation is the basis for recent work on inductive policy selection in relational planning domains, both deterministic (Khardon, 1999a; Martin & Geffner, 2000), and probabilistic (Yoon, Fern, & Givan, 2002). These techniques show that useful policies can be learned using a policy-space bias described by a generic (relational) knowledge representation language. Here we incorporate those ideas into a variant of API, that achieves significant success without representing or learning approximate value functions. Of course, a natural direction for future work is to combine policy-space techniques with value-function techniques, to leverage the advantages of both.

Given an initial policy, our approach uses the simulation technique of policy rollout (Tesauro & Galperin, 1996) to generate trajectories of an improved policy. These trajectories are then given to a classification learner, which searches for a classifier, or policy, that "matches" the trajectory data, resulting in an approximately improved policy. These two steps are iterated until no further improvement is observed. The resulting algorithm can be viewed as a form of API where the iteration is carried out without inducing approximate value functions.

---

1. Recent work in *relational reinforcement learning* has been applied to STRIPS problems with much simpler goals than typical benchmark planning domains, and is discussed in Section 8.

By avoiding value function learning, this algorithm helps address the representational challenge of applying API to relational planning domains. However, another fundamental challenge is that, for non-trivial relational domains, API requires some form of bootstrapping. In particular, for most STRIPS planning domains the reward, which corresponds to achieving a goal condition, is sparsely distributed and unlikely to be reached by random exploration. Thus, initializing API with a random or uninformed policy, will likely result in no reward signal and hence no guidance for policy improvement. One approach to bootstrapping is to rely on the user to provide a good initial policy or heuristic that gives guidance toward achieving reward. Rather, in this work we develop a new automatic bootstrapping approach for goal-based planning domains, which does not require user intervention.

Our bootstrapping approach is based on the idea of random-walk problem distributions. For a given planning domain, such as the blocks world, this distribution randomly generates a problem (i.e., an initial state and a goal) by selecting a random initial state and then executing a sequence of $n$ random actions, taking the goal condition to be a subset of properties from the resulting state. The problem difficulty typically increases with $n$, and for small $n$ (short random walks) even random policies can uncover reward. Intuitively, a good policy for problems with walk length $n$ can be used to bootstrap API for problems with slightly longer walk lengths. Our bootstrapping approach iterates this idea, by starting with a random policy and very small $n$, and then gradually increasing the walk length until we learn a policy for very long random walks. Such long-random-walk policies clearly capture much domain knowledge, and can be used in various ways. Here, we show that empirically such policies often perform well on problem distributions from relational domains used in recent deterministic and probabilistic planning competitions.

An implementation of this bootstrapped API approach took second place of 3 competitors in the hand-tailored track of the 2004 International Probabilistic Planning Competition.[2] To our knowledge this is the first machine-learning based system to be entered in any planning competition, either deterministic or probabilistic.

Here, we give an evaluation of our system on a number of probabilistic and deterministic relational planning domains, including the AIPS-2000 competition benchmarks, and benchmarks from the hand-tailored track of the 2004 Probabilistic Planning Competition. The results show that the system is often able to learn policies in these domains that perform well for long-random-walk problems. In addition, these same policies often perform well on the planning-competition problem distributions, comparing favorably with the state-of-the-art planner FF in the deterministic domains. Our experiments also highlight a number of limitations of the current system, which point to interesting directions for future work.

The remainder of paper proceeds as follows. In Section 2, we introduce our problem setup and then, in Section 3, present our new variant of API. In Section 4, we provide some technical analysis of the algorithm, giving performance bounds on the policy-improvement step. In Sections 5 and 6, we describe an implemented instantiation of our API approach for relational planning domains. This includes a description of a generic policy language for relational domains, a classification learner for that language, and a novel bootstrapping technique for goal-based domains. Section 7 presents our empirical results, and finally Sections 8 and 9 discuss related work and future directions.

---

2. Note, however, that this approach is not hand-tailored. Rather, given a domain definition, our system learns a policy offline, automatically, which can then be applied to any problem from the domain. We entered the hand-tailored track because it was the only track that facilitated the use of offline learning, by providing domains and problem generators before the competition. The other entrants were human-written for each domain.

## 2. Problem Setup

We formulate our work in the framework of Markov Decision Processes (MDPs). While our primary motivation is to develop algorithms for relational planning domains, we first describe our problem setup and approach for a general, action-simulator–based MDP representation. Later, in Section 5, we describe a particular representation of planning domains as relational MDPs and the corresponding relational instantiation of our approach.

Following and adapting Kearns, Mansour, and Ng (2002) and Bertsekas and Tsitsiklis (1996), we represent an MDP using a generative model $\langle S, A, T, R, I \rangle$, where $S$ is a finite set of states, $A$ is a finite, *ordered* set of actions, and $T$ is a randomized action-simulation algorithm that, given state $s$ and action $a$, returns a next state $s'$ according to some unknown probability distribution $P_T(s'|s, a)$. The component $R$ is a reward function that maps $S \times A$ to real-numbers, with $R(s, a)$ representing the reward for taking action $a$ in state $s$, and $I$ is a randomized initial-state algorithm with no inputs that returns a state $s$ according to some unknown distribution $P_0(s)$. We sometimes treat $I$ and $T(s, a)$ as random variables with distributions $P_0(\cdot)$ and $P_T(\cdot|s, a)$ respectively.

For an MDP $M = \langle S, A, T, R, I \rangle$, a policy $\pi$ is a (possibly stochastic) mapping from $S$ to $A$. The *value function* of $\pi$, denoted $V^\pi(s)$, represents the expected, cumulative, discounted reward of following policy $\pi$ in $M$ starting from state $s$, and is the unique solution to

$$V^\pi(s) = E[R(s, \pi(s)) + \gamma V^\pi(T(s, \pi(s)))] \tag{1}$$

where $0 \leq \gamma < 1$ is the discount factor. The *Q-value function* $Q^\pi(s, a)$ represents the expected, cumulative, discounted reward of taking action $a$ in state $s$ and then following $\pi$, and is given by

$$Q^\pi(s, a) = R(s, a) + \gamma E[V^\pi(T(s, a))] \tag{2}$$

We will measure the quality of a policy by the objective function $\overline{V}(\pi) = E[V^\pi(I)]$, giving the expected value obtained by that policy when starting from a randomly drawn initial state. A common objective in MDP planning and reinforcement learning is to find an optimal policy $\pi^* = \mathrm{argmax}_\pi \overline{V}(\pi)$. However, no automated technique, including the one we present here, has to date been able to guarantee finding an optimal policy in the relational planning domains we consider, in reasonable running time.

It is a well known fact that given a current policy $\pi$, we can define a new improved policy

$$\mathcal{P}I^\pi(s) = \mathrm{argmax}_{a \in A} Q^\pi(s, a) \tag{3}$$

such that the value function of $\mathcal{P}I^\pi$ is guaranteed to 1) be no worse than that of $\pi$ at each state $s$, and 2) strictly improve at some state when $\pi$ is not optimal. *Policy iteration* is an algorithm for computing optimal policies by iterating policy improvement ($\mathcal{P}I$) from any initial policy to reach a fixed point, which is guaranteed to be an optimal policy. Each iteration of policy improvement involves two steps: 1) *Policy evaluation* where we compute the value function $V^\pi$ of the current policy $\pi$, and 2) *Policy selection*, where, given $V^\pi$ from step 1, we select the action that maximizes $Q^\pi(s, a)$ at each state, defining a new improved policy.

**Finite Horizons.** Since our API variant is based on simulation, and must bound the simulation trajectories by a horizon $h$, our technical analysis in Section 4 will use the notion of finite-horizon discounted reward. The $h$-horizon value function $V_h^\pi$ is recursively defined as

$$V_0^\pi(s) = 0, \quad V_h^\pi(s) = E[R(s, \pi(s)) + \gamma V_{h-1}(T(s, \pi(s)))] \tag{4}$$

giving the expected discounted reward obtained by following $\pi$ for $h$ steps from $s$. We also define the $h$-horizon Q-function $Q_h^\pi(s, a) = R(s, a) + \gamma E[V_{h-1}^\pi(T(s, a))]$, and the $h$-horizon objective function $\overline{V}_h(\pi) = E[V_h^\pi(I)]$. It is well known, that the effect of using a finite horizon can be made arbitrarily small. In particular, we have that for all states $s$ and actions $a$, the approximation error decreases exponentially with $h$,

$$
\begin{aligned}
|V^\pi(s) - V_h^\pi(s)| &\leq \gamma^h V_{\max}, \\
|Q^\pi(s, a) - Q_h^\pi(s, a)| &\leq \gamma^h V_{\max}, \text{ and} \\
V_{\max} &= \frac{R_{\max}}{1 - \gamma},
\end{aligned}
$$

where $R_{\max}$ is the maximum of the absolute value of the reward for any action at any state. From this we also get that $|\overline{V}_h(\pi) - \overline{V}(\pi)| \leq \gamma^h V_{\max}$.

## 3. Approximate Policy Iteration with a Policy Language Bias

Exact solution techniques, such as policy iteration, are typically intractable for large state-space MDPs, such as those arising from relational planning domains. In this section, we introduce a new variant of approximate policy iteration (API) intended for such domains. First, we review a generic form of API used in prior work, based on learning approximate value functions. Next, motivated by the fact that value functions are often difficult to learn in relational domains, we describe our API variant, which avoids learning value functions and instead learns policies directly as state-action mappings.

### 3.1 API with Approximate Value Functions

API, as described in Bertsekas and Tsitsiklis (1996), uses a combination of Monte-Carlo simulation and inductive machine learning to heuristically approximate policy iteration in large state-space MDPs. Given a current policy $\pi$, each iteration of API approximates policy evaluation and policy selection, resulting in an approximately improved policy $\hat{\pi}$. First, the policy-evaluation step constructs a training set of samples of $V^\pi$ from a small but representative set of states. Each sample is computed using simulation, estimating $V^\pi(s)$ for the policy $\pi$ at each state $s$ by drawing some number of sample trajectories of $\pi$ starting at $s$ and then averaging the cumulative, discounted reward along those trajectories. Next, the policy-selection step uses a function approximator (e.g., a neural network) to learn an approximation $\hat{V}^\pi$ to $V^\pi$ based on the training data. $\hat{V}^\pi$ then serves as a representation for $\hat{\pi}$, which selects actions using sampled one-step lookahead based on $\hat{V}^\pi$, that is

$$\hat{\pi}(s) = \arg\max_{a \in A} R(s, a) + \gamma E[\hat{V}^\pi(T(s, a))].$$

A common variant of this procedure learns an approximation of $Q^\pi$ rather than $V^\pi$.

API exploits the function approximator's generalization ability to avoid evaluating each state in the state space, instead only directly evaluating a small number of training states. Thus, the use of API assumes that states and perhaps actions are represented in a factored form (typically, a feature vector) that facilitates generalizing properties of the training data to the entire state and action spaces. Note that in the case of perfect generalization (i.e., $\hat{V}^\pi(s) = V^\pi(s)$ for all states $s$), we have that $\hat{\pi}$ is equal to the exact policy improvement $\mathcal{PI}^\pi$, and thus API simulates exact policy iteration. However, in practice, generalization is not perfect, and there are typically no guarantees for policy improvement[3]—nevertheless, API often "converges" usefully (Tesauro, 1992; Tsitsiklis & Van Roy, 1996).

The success of the above API procedure depends critically on the ability to represent and learn good value-function approximations. For some MDPs, such as those arising from relational planning domains, it is often difficult to specify a space of value functions and learning mechanism that facilitate good generalization. For example, work in relational reinforcement learning (Dzeroski, DeRaedt, & Driessens, 2001) has shown that learning approximate value functions for classical domains, such as the blocks world, can be problematic.[4] In spite of this, it is often relatively easy to compactly specify good policies using a language for (relational) state-action mappings. This suggests that such languages may provide useful policy-space biases for learning in API. However, all prior API methods are based on approximating value functions and hence can not leverage these biases. With this motivation, we consider a form of API that directly learns policies without directly representing or approximating value functions.

## 3.2 Using a Policy Language Bias

A policy is simply a classifier, possibly stochastic, that maps states to actions. Our API approach is based on this view, and is motived by recent work that casts policy selection as a standard classification learning problem. In particular, given the ability to observe trajectories of a target policy, we can use machine learning to select a policy, or classifier, that mimics the target as closely as possible. Khardon (1999b) studied this learning setting and provided PAC-like learnability results, showing that under certain assumptions, a small number of trajectories is sufficient to learn a policy whose value is close to that of the target. In addition, recent empirical work, in relational planning domains (Khardon, 1999a; Martin & Geffner, 2000; Yoon et al., 2002), has shown that by using expressive languages for specifying state-action mappings, good policies can be learned from sample trajectories of good policies.

---

3. Under very strong assumptions, API can be shown to converge in the infinite limit to a near-optimal value function. See Proposition 6.2 of Bertsekas and Tsitsiklis (1996).

4. In particular, the RRL work has considered a variety of value-function representation including relational regression trees, instance based methods, and graph kernels, but none of them have generalized well over varying numbers of objects.

These results suggest that, given a policy $\pi$, if we can somehow generate trajectories of an improved policy, then we can learn an approximately improved policy based on those trajectories. This idea is the basis of our approach. Figure 1 gives pseudo-code for our API variant, which starts with an initial policy $\pi_0$ and produces a sequence of approximately improved policies. Each iteration involves two primary steps: First, given the current policy $\pi$, the procedure **Improved-Trajectories** (approximately) generates trajectories of the improved policy $\pi' = \mathcal{P}I^\pi$. Second, these trajectories are used as training data for the procedure **Learn-Policy**, which returns an approximation of $\pi'$. We now describe each step in more detail.

**Step 1: Generating Improved Trajectories.** Given a base policy $\pi$, the simulation technique of *policy rollout* (Tesauro & Galperin, 1996; Bertsekas & Tsitsiklis, 1996) computes an approximation $\hat{\pi}$ to the improved policy $\pi' = \mathcal{P}I^\pi$, where $\pi'$ is the result of applying one step of policy iteration to $\pi$. Furthermore, for a given state $s$, policy rollout computes $\hat{\pi}(s)$ without the need to solve for $\pi'$ at all other states, and thus provides a tractable way to approximately simulate the improved policy $\pi'$ in large state-space MDPs. Often $\pi'$ is significantly better than $\pi$, and hence so is $\hat{\pi}$, which can lead to substantially improved performance at a small cost. Policy rollout has provided significant benefits in a number of application domains, including for example Backgammon (Tesauro & Galperin, 1996), instruction scheduling (McGovern, Moss, & Barto, 2002), network-congestion control (Wu, Chong, & Givan, 2001), and Solitaire (Yan, Diaconis, Rusmevichientong, & Van Roy, 2004).

Policy rollout computes $\hat{\pi}(s)$, the estimate of $\pi'(s)$, by estimating $Q^\pi(s, a)$ for each action $a$ and then taking the maximizing action to be $\hat{\pi}(s)$ as suggested by Equation 3. Each $Q^\pi(s, a)$ is estimated by drawing $w$ trajectories of length $h$, where each trajectory is the result of starting at $s$, taking action $a$, and then following the actions selected by $\pi$ for $h - 1$ steps. The estimate of $Q^\pi(s, a)$ is then taken to be the average of the cumulative discounted reward along each trajectory. The *sampling width $w$* and *horizon $h$* are specified by the user, and control the trade off between increased computation time for large values, and reduced accuracy for small values. Note that rollout applies to both stochastic and deterministic policies and that due to variance in the Q-value estimates, the rollout policy can be stochastic even for deterministic base policies.

The procedure **Improved-Trajectories** uses rollout to generate $n$ length $h$ trajectories of $\hat{\pi}$, each beginning at a randomly drawn initial state. Rather than just recording the states and actions along each trajectory, we store additional information that is used by our policy-learning algorithm. In particular, the i'th element of a trajectory has the form $\langle s_i, \pi(s_i), \hat{Q}(s_i, a_1), \ldots, \hat{Q}(s_i, a_m) \rangle$, giving the i'th state $s_i$ along the trajectory, the action selected by the current (unimproved) policy at $s_i$, and the Q-value estimates $\hat{Q}(s_i, a)$ for each action. Note that given the Q-value information for $s_i$ the learning algorithm can determine the approximately improved action $\hat{\pi}(s)$, by maximizing over actions, if desired.

**Step 2: Learn Policy.** Intuitively, we want **Learn-Policy** to select a new policy that closely matches the training trajectories. In our experiments, we use relatively simple learning algorithms based on greedy search within a space of policies specified by a policy-language bias. In Sections 5.2 and 5.3 we detail the policy-language learning bias used by our technique, and the associated learning algorithm. In Section 4 we provide some technical analysis of an idealized version of this algorithm, providing guidance regarding the required number of training trajectories. We note that by labeling each training state in the trajectories with the associated $Q$-values for each action, rather than simply with the best action, we enable the learner to make more informed trade-offs, focusing on accuracy at states where wrong decisions have high costs, which was empirically useful. Also, the inclusion of $\pi(s)$ in the training data enables the learner to adjust the data relative to $\pi$, if desired—e.g., our learner uses a bias that focuses on states where large improvement appears possible.

Finally, we note that for API to be effective, it is important that the initial policy $\pi_0$ provide guidance toward improvement, i.e., $\pi_0$ must bootstrap the API process. For example, in goal-based planning domains $\pi_0$ should reach a goal from some of the sampled states. In Section 6 we will discuss this important issue of bootstrapping and introduce a new bootstrapping technique.

## 4. Technical Analysis

In this section, we consider a variant of the policy improvement step of our main API loop, which learns an improved policy given a base policy $\pi$. We show how to select a sampling width $w$, horizon $h$, and training set size $n$ such that, under certain assumptions, the quality of the learned policy is close to the quality of $\pi'$ the policy iteration improvement. Similar results have been shown for previous forms of API based on approximate value functions (Bertsekas & Tsitsiklis, 1996), however, our assumptions are of a much different nature.[5]

The analysis is divided into two parts. First, following Khardon (1999b), we consider the sample complexity of policy learning. That is, we consider how many trajectories of a target policy must be observed by a learner before we can guarantee a good approximation to the target. Second, we show how to apply this result, which is for deterministic policies, to the problem of learning from rollout policies, which can be stochastic. Throughout we assume the context of an MDP $M = \langle S, A, T, R, I \rangle$.

### 4.1 Learning Deterministic Policies

A *trajectory* of length $h$ is a sequence $(s_0, a_0, s_1, a_1, \ldots, a_{h-1}, s_h)$ of alternating states $s_i$ and actions $a_i$. We say that a *deterministic* policy $\pi$ is *consistent* with a trajectory $(s_1, a_1, \ldots, s_h)$ if and only if for each $0 \leq i < h$, $\pi(s_i) = a_i$. We define $D_h^\pi$ to be a distribution over the set of all length $h$ trajectories, such that $D_h^\pi(t)$ is the probability that $\pi$ generates trajectory $t = (s_0, a_0, s_1, a_1, \ldots, a_{h-1}, s_h)$ according to the following process: first draw $s_0$ according to the initial state distribution $I$, and then draw $s_{i+1}$ from $T(s_i, \pi(s_i))$ for $0 \leq i < h$. Note that $D_h^\pi(t)$ is non-zero only if $\pi$ is consistent with $t$.

Our policy improvement step first generates trajectories of the rollout policy $\hat{\pi}$ (see Section 3.2), via the procedure **Improved-Trajectories**, and then learns an approximation

---

5. In particular, Bertsekas and Tsitsiklis (1996) assumes a bound on the $L_\infty$ norm of the value function approximation, i.e., that at each state the approximation is almost perfect. Rather we assume that the improved policy $\pi'$ comes from a finite class of policies for which we have a consistent learner. In both cases policy improvement can be guaranteed given an additional assumption on the minimum $Q$-advantage of the MDP (see below).

---

**API** $(n, w, h, M, \pi_0, \gamma)$

// training set size $n$, sampling width $w$, horizon $h$,
// MDP $M = \langle S, \{a_1, \ldots, a_m\}, T, R, I \rangle$, initial policy $\pi_0$, discount factor $\gamma$.

$\pi \leftarrow \pi_0$;
**loop**
    $T \leftarrow$ **Improved-Trajectories**$(n, w, h, M, \pi)$;
    $\pi \leftarrow$ **Learn-Policy**$(T)$;
**until** satisfied with $\pi$;   // e.g., until change is small
**Return** $\pi$;

---

**Improved-Trajectories**$(n, w, h, M, \pi)$

// training set size $n$, sampling width $w$,
// horizon $h$, MDP $M$, current policy $\pi$

$T \leftarrow \emptyset$;
**repeat** $n$ times   // generate $n$ trajectories of improved policy
    $t \leftarrow$ **nil**;
    $s \leftarrow$ state drawn from $I$;  // draw random initial state
    **for** $i = 1$ to $h$
        $\langle \hat{Q}(s, a_1), \ldots, \hat{Q}(s, a_m) \rangle \leftarrow$ **Policy-Rollout**$(\pi, s, w, h, M)$;  // $Q^\pi(s, a)$ estimates
        $t \leftarrow t \cdot \langle s, \pi(s), \hat{Q}(s, a_1), \ldots, \hat{Q}(s, a_m) \rangle \rangle$;  // concatenate new sample onto trajectory
        $a \leftarrow$ action maximizing $\hat{Q}(s, a)$;  // action of the improved policy at state $s$
        $s \leftarrow$ state sampled from $T(s, a)$;  // simulate action of improved policy
    $T \leftarrow T \cup t$;
**Return** $T$;

---

**Policy-Rollout** $(\pi, s, w, h, M)$   // Compute $Q^\pi(s, a)$ estimates $\langle \hat{Q}(s, a_1), \ldots, \hat{Q}(s, a_m) \rangle$

// policy $\pi$, state $s$, sampling width $w$, horizon $h$, MDP $M$

**for** each action $a_i$ in $A$
    $\hat{Q}(s, a_i) \leftarrow 0$;
    **repeat** $w$ times   // $\hat{Q}(s, a_i)$ is an average over $w$ trajectories
        $R \leftarrow R(s, a_i)$;  $s' \leftarrow$ a state sampled from $T(s, a_i)$;  // take action $a_i$ in $s$
        **for** $i = 1$ to $h - 1$  // take $h - 1$ steps of $\pi$ accumulating discounted reward in R
            $R \leftarrow R + \gamma^i R(s', \pi(s'))$;
            $s' \leftarrow$ a state sampled from $T(s', \pi(s'))$
        $\hat{Q}(s, a_i) \leftarrow \hat{Q}(s, a_i) + R$;  // include trajectory in average
    $\hat{Q}(s, a_i) \leftarrow \frac{\hat{Q}(s, a_i)}{w}$;
**Return** $\langle \hat{Q}(s, a_1), \ldots, \hat{Q}(s, a_m) \rangle$

---

Figure 1: Pseudo-code for our API algorithm. See Section 5.3 for an instantiation of **Learn-Policy** called **Learn-Decision-List**.

of $\hat{\pi}$. Note that the rollout policy serves as a stochastic approximation of $\pi' = \mathcal{P}I^\pi$ the policy iteration improvement of $\pi$. Thus, **Improved-Trajectories** can be viewed as attempting to draw trajectories from $D_h^{\pi'}$, and the learning step can be viewed as learning an approximation of $\pi'$. Imagining for the moment that we can draw trajectories from $D_h^{\pi'}$, a fundamental question is how many trajectories are sufficient to ensure that the learned policy will be about as good as $\pi'$. Khardon (1999b) studied this question for the case of deterministic policies in undiscounted goal-based planning domains (i.e., MDPs where reward is only received at goal states). Here we give a straightforward adaptation of his main result to our problem setting where we have general reward functions and measure the quality of a policy by $\overline{V}(\pi)$.

The learning-problem formulation is similar in spirit to the standard framework of Probably Approximately Correct (PAC) learning. In particular, we will assume that the target policy comes from a finite class of deterministic policies $H$. For example, $H$ may correspond to the set of policies that can be described by bounded-length decision lists. In addition, we assume that the learner is consistent—i.e., it returns a policy from $H$ that is consistent with all of the training trajectories. Under these assumptions, a relatively small number of trajectories (logarithmic in $|H|$) are sufficient to ensure that with high probability the learned policy is about as good as the target.

**Proposition 1.** *Let $H$ be a finite class of deterministic policies. For any $\pi \in H$, and any set of $n = \epsilon^{-1} \ln \frac{|H|}{\delta}$ trajectories drawn independently from $D_h^\pi$, there is a $1 - \delta$ probability that every $\hat{\pi} \in H$ consistent with the trajectories satisfies $\overline{V}(\hat{\pi}) \geq \overline{V}(\pi) - 2V_{max}(\epsilon + \gamma^h)$.*

The proof of this proposition is in the Appendix. The computational complexity of finding a consistent policy depends on the policy class $H$. Polynomial-time algorithms can be given for interesting classes such as bounded-length decision lists—however, these algorithms are typically too expensive for the policy classes we consider in practice. Rather, as described in Section 5.3, we use a learner based on greedy heuristic search, which often works well in practice.

The assumption that the target policy comes from a fixed size class $H$ will often be violated. However, as pointed out by Khardon (1999b), it is straightforward to give an extension of Proposition 1 for the setting where the learner considers increasingly complex policies until a consistent one is found. In this case, the sample complexity is related to the encoding size of the target policy rather than the size of $H$, thus allowing the use of very large and expressive policy classes without necessarily paying the full sample-complexity price of Proposition 1.

### 4.2 Learning from Rollout Policies

The proof of Proposition 1 relies critically on the fact that the policy class $H$ contains only deterministic policies. However, in our main API loop, the target policies are computed via rollout and hence are stochastic due to the uncertainty introduced by finite sampling. Thus, we cannot directly use Proposition 1 in the context of learning from trajectories produced by rollout. To deal with this problem we describe a variant of **Improved-Trajectories** that can reliably generate training trajectories from the deterministic policy $\pi' = \mathcal{P}I^\pi$ (see Equation 3), which is guaranteed to improve on $\pi$ if improvement is possible.

Given a base policy $\pi$, we first define $A^\pi(s)$ to be the set of actions that maximize $Q^\pi(s, a)$. Note that $\pi'(s) = \min A^\pi(s)$, where the minimum is taken with respect to the action ordering provided by the MDP. Importantly this policy is deterministic and thus if we can generate trajectories of it, then we can apply the above result to learn a close approximation. In order to generate trajectories of $\pi'$ we slightly modify **Improved-Trajectories**. The modification is introduced for analysis only, and our experiments are based on the procedures given in Figure 1. Our modification is to replace the action maximization step of **Improved-Trajectories** (second to last statement of the for loop), which chooses the next action $a$ to execute, with the following two steps

$$\hat{A}(\Delta, s) \quad \leftarrow \quad \{a' \mid \max_a \hat{Q}(s, a) - \hat{Q}(s, a') \leq \Delta\}$$
$$a \quad \leftarrow \quad \min \hat{A}(\Delta, s)$$

where $\hat{Q}(s, a)$ is the estimate of $Q_h^\pi(s, a)$ computed by policy rollout using a sampling width $w$, and $\Delta$ is a newly introduced parameter.

Note that if $\hat{A}(\Delta, s) = A^\pi(s)$, then the selected action $a$ will equal $\pi'(s)$. If this condition is true for every state encountered then the modified **Improved-Trajectories** will effectively generate trajectories of $\pi'$. Thus, we would like to bound the probability that $\hat{A}(\Delta, s) \neq A^\pi(s)$ to a small value by appropriately choosing the sampling width $w$, the horizon $h$, and $\Delta$. Unfortunately, the choice of these parameters depends on the MDP. That is, given any particular parameter values, there is an MDP such that the event $\hat{A}(\Delta, s) \neq A^\pi(s)$ has a non-negligible probability at some state. For this reason we first define the *Q-advantage* $\Delta^*$ of an MDP and show how to select appropriate parameter values given a lower-bound on $\Delta^*$.

Given an MDP and policy $\pi$, let $S'$ be the set of states such that $s \in S'$ iff there are two actions $a$ and $a'$ such that $Q^\pi(s, a) \neq Q^\pi(s, a')$, i.e., there are actions with distinct Q-values. Also for each state in $S'$ define $a_1^*(s)$ and $a_2^*(s)$ be a best action and a second best action respectively as measured by $Q^\pi(s, a)$. The Q-advantage is defined as $\Delta^* = \min_{s \in S'} a_1^*(s) - a_2^*(s)$, which measures the minimum Q-value gap between an optimal and sub-optimal action over the state space. Given a lower-bound on the Q-advantage of an MDP the following proposition indicates how to select parameter values to ensure that $\hat{A}(\Delta, s) = A^\pi(s)$ with high probability.

**Proposition 2.** *For any MDP with Q-advantage at least $\Delta^*$, and any $0 < \delta' < 1$, if we have*

$$h \quad > \quad \log_\gamma \frac{\Delta^*}{8V_{max}}$$
$$w \quad > \quad \left(\frac{8V_{max}}{\Delta^*}\right)^2 \ln \frac{|A|}{\delta'}$$
$$\Delta \quad = \quad \frac{\Delta^*}{2}$$

*then for any state $s$, $\hat{A}(\Delta, s) = A^\pi(s)$ with probability at least $1 - \delta'$.*

The proof is given in the Appendix. Thus, for parameter values satisfying the above conditions, if our MDP has Q-advantage at least $\Delta^*$ then we are guaranteed that with probability at least $1 - \delta'$ that $\hat{A}(\Delta, s) = A(\Delta, s)$. This means that **Improved-Trajectories** will correctly select the action $\pi'(s)$ with probability at least $1 - \delta'$. Note that this proposition agrees with the intuition that both $h$ and $w$ should increase with decreasing Q-advantage and increasing $V_{\max}$—and also that $w$ should increase for decreasing $\delta'$.[6]

In order to generate $n$ length $h$ trajectories of $\pi'$, the modified **Improved-Trajectories** routine must compute the set $\hat{A}(\Delta, \cdot)$ at $n \cdot h$ states, yielding $n \cdot h$ opportunities to make an error. To ensure that no error is made, the modified procedure sets the sampling width $w$ using $\delta' = \frac{\delta}{2nh}$. This guarantees that an error free training set is created with probability at least $1 - \frac{\delta}{2}$.

Combining this observation with the assumption that $\pi' \in H$ we can apply Proposition 1 as follows. First, generate $n = \epsilon^{-1} \ln \frac{2|H|}{\delta}$ trajectories of $\pi'$ using the modified **Improved-Trajectories** routine (with $\delta' = \frac{\delta}{2nh}$). Next, learn a policy $\hat{\pi}$ from these trajectories using a consistent learner. We know that the probability of generating an imperfect training set is bounded by $\frac{\delta}{2}$, and for the chosen value of $n$, the failure probability of the learner is also bounded by $\frac{\delta}{2}$. Thus, we get that with probability at least $1 - \delta$, the learned policy $\hat{\pi}$ satisfies $\overline{V}(\hat{\pi}) \geq \overline{V}(\pi') - 2V_{\max}(\epsilon + \gamma^h)$, giving an approximation guarantee relative to the improved policy $\pi'$. This is summarized by the following proposition.

**Proposition 3.** *Let $H$ be a finite class of deterministic policies, $0 < \delta < 1$, and $0 < \epsilon < 1$. For any MDP with Q-advantage at least $\Delta^*$, any policy $\pi$ such that $\mathcal{P}I^\pi \in H$, and any set of $n > \epsilon^{-1} \ln\left(2|H|\delta^{-1}\right)$ trajectories produced by modified* **Improved-Trajectories** *using parameters satisfying,*

$$\Delta = \frac{\Delta^*}{2}$$
$$h > \log_\gamma \frac{\Delta^*}{8V_{max}}$$
$$w > \left(\frac{8V_{max}}{\Delta^*}\right)^2 \ln \frac{2nh|A|}{\delta}$$

*there is at least a $1 - \delta$ probability that every $\hat{\pi} \in H$ consistent with the trajectories satisfies $\overline{V}(\hat{\pi}) \geq \overline{V}(\mathcal{P}I^\pi) - 2V_{max}(\epsilon + \gamma^h)$.*

One notable aspect of this result is that there is only a logarithmic dependence on the number of actions $|A|$ and $\delta^{-1}$. However, the practical utility is hindered by its dependence on $\Delta^*$ which is typically not known in practice, and can be exponentially small in the planning horizon. Unfortunately, this dependence appears to be unavoidable for our type of approach where we try to learn from trajectories of $\mathcal{P}I^\pi$ produced by rollout. This is because for any particular setting of the above parameters, there is always an MDP with a small enough Q-advantage, such that the value of the rollout policy is arbitrarily worse than that of $\mathcal{P}I^\pi$.

## 5. API for Relational Planning

Our work is motivated by the goal of solving relational MDPs. In particular, we are interested in finding policies for relational MDPs that represent classical planning domains and their stochastic variants. Such policies can then be applied to any problem instance from a planning domain, and hence can be viewed as a form of domain-specific control knowledge.

---

6. At first glance it appears that the lower-bound on $h$ decreases with increasing $V_{\max}$ and decreasing $\Delta^*$. However, the opposite is true since the base of the logarithm is the discount factor, which is strictly less than one. Also note that since $\Delta^*$ is upper-bounded by $2V_{\max}$ the bound on $h$ will always be positive.

In this section, we first describe a straightforward way to view classical planning domains (not just single problem instances) as relationally factored MDPs. Next, we describe our relational policy space in which policies are compactly represented as taxonomic decision lists. Finally, we present a heuristic learning algorithm for this policy space.

## 5.1 Planning Domains as MDPs.

We say that an MDP $\langle S, A, T, R, I \rangle$ is *relational* when $S$ and $A$ are defined by giving a finite set of objects $O$, a finite set of predicates $P$, and a finite set of action types $Y$. A *fact* is a predicate applied to the appropriate number of objects, e.g., $\mathbf{on}(a, b)$ is a blocks-world fact. A state is a set of facts, interpreted as representing the true facts in the state. The state space $S$ contains all possible sets of facts. An *action* is an action type applied to the appropriate number of objects, e.g., $\mathbf{putdown}(a)$ is a blocks-world action, and the action space $A$ is the set of all such actions.

A classical planning domain describes a set of problem instances with related structure, where a problem instance gives an initial world state and goal. For example, the blocks world is a classical planning domain, where each problem instance specifies an initial block configuration and a set of goal conditions. Classical planners attempt to find solutions to specific problem instances of a domain. Rather, our goal is to solve entire planning domains by finding a policy that can be applied to all problem instances. As described below, it is straightforward to view a classical planning domain as a relational MDP where each MDP state corresponds to a problem instance.

**State and Action Spaces.** Each classical planning domain specifies a set of action types $Y$, *world predicates* $W$, and possible world objects $O$. Together $Y$ and $O$ define the MDP action space. Each state of the MDP corresponds to a single problem instance (i.e., a world state and a goal) from the planning domain by specifying both the current world and the goal. We achieve this by letting the set of relational MDP predicates be $P = W \cup G$, where $G$ is a set of *goal predicates*. The set of goal predicates contains a predicate for each world predicate in $W$, which is named by prepending a 'g' onto the corresponding world predicate name (e.g., the goal predicate $\mathbf{gclear}$ corresponds to the world predicate $\mathbf{clear}$). With this definition of $P$ we see that the MDP states are sets of goal and world facts, indicating the true world facts of a problem instance and the goal conditions. It is important to note, as described below, that the MDP actions will only change world facts and not goal facts. Thus, this large relational MDP can be viewed as a collection of disconnected sub-MDPs, where each sub-MDP corresponds to a distinct goal condition.

**Reward Function.** Given an MDP state the objective is to reach another MDP state where the goal facts are a subset of the corresponding world facts—i.e., reach a world state that satisfies the goal. We will call such states *goal states* of the MDP. For example, the MDP state

$$\{\mathbf{on\text{-}table}(a), \mathbf{on}(a, b), \mathbf{clear}(b), \mathbf{gclear}(b)\}$$

is a goal state in a blocks-world MDP, but would not be a goal state without the world fact $\mathbf{clear}(b)$. We represent the objective of reaching a goal state quickly by defining $R$ to assign a reward of zero for actions taken in goal states and negative rewards for actions in all other states, representing the cost of taking those actions. Typically, for classical planning domains, the action costs are uniformly -1, however, our framework allows the cost to vary across actions.

**Transition Function.** Each classical planning domain provides an action simulator (e.g., as defined by STRIPS rules) that, given a world state and action, returns a new world state. We define the MDP transition function $T$ to be this simulator modified to treat goal states as terminal and to preserve without change all goal predicates in an MDP state. Since classical planning domains typically have a large number of actions, the action definitions are usually accompanied by preconditions that indicate the *legal actions* in a given state, where usually the legal actions are a small subset of all possible actions. We assume that $T$ treats actions that are not legal as no-ops. For simplicity, our relational MDP definition does not explicitly represent action preconditions, however, we assume that our algorithms do have access to preconditions and thus only need to consider legal actions. For example, we can restrict rollout to only the legal actions in a given state.

**Initial State Distribution.** Finally, the initial state distribution $I$ can be any program that generates legal problem instances (MDP states) of the planning domain. For example, problem domains from planning competitions are commonly distributed with problem generators.

With these definitions, a good policy is one that can reach goal states via low-cost action sequences from initial states drawn from $I$. Note that here policies are mappings from problem instances to actions and thus can be sensitive to goal conditions. In this way, our learned policies are able to generalize across different goals. We next describe a language for representing such generalized policies.

## 5.2 Taxonomic Decision List Policies.

For single argument action types, many useful rules for planning domains take the form of "apply action type $A$ to any object in class $C$" (Martin & Geffner, 2000). For example, in the blocks world, "pick up any clear block that belongs on the table but is not on the table", or in a logistics world, "unload any object that is at its destination". Using a concept language for describing object classes, Martin and Geffner (2000) introduced the use of decision lists of such rules as a useful learning bias, showing promising experiments in the deterministic blocks world. With that motivation, we consider a policy space that is similar to the one used originally by Martin and Geffner, but generalized to handle multiple action arguments. Also, for historical reasons, our concept language is based upon taxonomic syntax (McAllester, 1991; McAllester & Givan, 1993), rather than on description logic as used by Martin and Geffner.

**Comparison Predicates.** For relational MDPs with world and goal predicates, such as those corresponding to classical planning domains, it is often useful for polices to compare the current state with the goal. To this end, we introduce a new set of predicates, called *comparison predicates*, which are derived from the world and goal predicates. For each world predicate $p$ and corresponding goal predicate $gp$, we introduce a new comparison predicate $cp$ that is defined as the conjunction of $p$ and $gp$. That is, a comparison-predicate fact is true if and only if both the corresponding world and goal predicates facts are true.

For example, in the blocks world, the comparison-predicate fact $\mathbf{con}(a, b)$ indicates that $a$ is on $b$ in both the current state and the goal—i.e., $\mathbf{on}(a, b)$ and $\mathbf{gon}(a, b)$ are true.

**Taxonomic Syntax.** Taxonomic syntax provides a language for writing class expressions that represent sets of objects with properties of interest and serve as the fundamental pieces with which we build policies. Class expressions are built from the MDP predicates (including comparison predicates if applicable) and variables. In our policy representation, the variables will be used to denote action arguments, and at runtime will be instantiated by objects. For simplicity we only consider predicates of arity one and two, which we call *primitive classes* and relations, respectively. When a domain contains predicates of arity three or more, we automatically convert them to multiple auxiliary binary predicates. Given a list of variables $X = (x_1, \ldots, x_k)$, class expressions are given by,

$$
\begin{aligned}
C[X] &::= C_0 \mid x_i \mid \mathbf{a\text{-}thing} \mid \neg C[X] \mid (R \; C[X]) \mid (\min \; R) \\
R &::= R_0 \mid R^{-1} \mid R^*
\end{aligned}
$$

where $C[X]$ is a class expression, $R$ is a relation expression, $C_0$ is a primitive class, $R_0$ is a primitive relation, and $x_i$ is a variable in $X$. Note that, for classical planning domains, the primitive classes and relations can be world, goal, or comparison predicates. We define the *depth* $d(C[X])$ of a class expression $C[X]$ to be one if $C[X]$ is either a primitive class, **a-thing**, a variable, or $(\min R)$, otherwise we define $d(\neg C[X])$ and $d(R \; C[X])$ to be $d(C[X]) + 1$, where $R$ is a relation expression and $C[X]$ is a class expression. For a given relational MDP we denote by $\mathcal{C}_d[X]$ the set of all class expressions $C[X]$ that have a depth of $d$ or less.

Intuitively the class expression $(R \; C[X])$ denotes the set of objects that are related through relation $R$ to some object in the set $C[X]$. The expression $(R^* \; C[X])$ denotes the set of objects that are related through some "$R$ chain" to an object in $C[X]$—this constructor is important for representing recursive concepts (e.g., the blocks above $a$). The expression $(\min R)$ denotes the set of objects that are minimal under the relation $R$.

More formally, let $s$ be an MDP state and $O = (o_1, \ldots, o_k)$ be a variable assignment, which assigns object $o_i$ to variable $x_i$. The interpretation of $C[X]$ relative to $s$ and $O$ is a set of objects and is denoted by $C[X]^{s,O}$. A primitive class $C_0$ is interpreted as the set of objects for which the predicate symbol $C_0$ is true in $s$. Likewise, a primitive relation $R_0$ is interpreted as the set of all object tuples for which the relation $R_0$ holds in $s$. The class expression **a-thing** denotes the set of all objects in $s$. The class expression $x_i$, where $x_i$ is a variable, is interpreted to be the singleton set $\{o_i\}$. The interpretation of compound expressions is given by,

$$
\begin{aligned}
(\neg C[X])^{s,O} &= \{o \mid o \notin C[X]^{s,O}\} \\
(R \, C[X])^{s,O} &= \{o \mid \exists o' \in C[X]^{s,O} \text{ s.t. } (o', o) \in R^{s,O}\} \\
(\min R)^{s,O} &= \{o \mid \exists o' \text{ s.t. } (o, o') \in R^{s,O}, \nexists o' \text{ s.t. } (o', o) \in R^{s,O}\} \\
(R^*)^{s,O} &= \mathbf{ID} \cup \{(o_1, o_v) \mid \exists o_2, \ldots, o_{v-1} \text{ s.t. } (o_i, o_{i+1}) \in R^{s,O} \text{ for } 1 \leq i < v\} \\
(R^{-1})^{s,O} &= \{(o, o') \mid (o', o) \in R^{s,O}\}
\end{aligned}
$$

where $C[X]$ is a class expression, $R$ is a relation expression, and $\mathbf{ID}$ is the identity relation. Some examples of useful blocks-world concepts, given the primitive classes **clear**, **gclear**, **holding**, and **con-table**, along with the primitive relations **on**, **gon**, and **con**, are:

- (**gon**$^{-1}$ **holding**) has depth two, and denotes the block that we want under the block being held.

- (**on**$^*$ (**on gclear**)) has depth three, and denotes the blocks currently above blocks that we want to make clear.

- (**con**$^*$ **con-table**) has depth two, and denotes the set of blocks in well constructed towers. To see this note that a block $b_v$ is in this class if and only if there exists a sequence of blocks $b_1, \ldots, b_v$ such that $b_1$ is on the table in both the goal and the current state (i.e. **con-table**$(b_1)$) and $b_{i+1}$ is on $b_i$ in both the goal and current state (i.e. **con**$(b_i, b_{i+1})$) for $1 \le i < v$.

- (**gon** (**con**$^*$ **con-table**)) has depth three, and denotes the blocks that belong on top of a currently well constructed tower.

**Decision List Policies** We represent policies as decision lists of *action-selection rules*. Each rule has the form $a(x_1, \ldots, x_k) : L_1, L_2, \ldots L_m$, where $a$ is a $k$-argument action type, the $L_i$ are *literals*, and the $x_i$ are action-argument variables. We will denote the list of action argument variables as $X = (x_1, \ldots, x_k)$. Each literal has the form $x \in C[X]$, where $C[X]$ is a taxonomic syntax class expression and $x$ is an action-argument variable.

Given an MDP state $s$ and a list of action-argument objects $O = (o_1, \ldots, o_k)$, we say that a literal $x_i \in C[X]$ is true given $s$ and $O$ iff $o_i \in C[X]^{s,O}$. We say that a rule $R = a(x_1, \ldots, x_k) : L_1, L_2, \ldots L_m$ *allows* action $a(o_1, \ldots o_k)$ in $s$ iff each literal in the rule is true given $s$ and $O$. Note that if there are no literals in a rule for action type $a$, then all possible actions of type $a$ are allowed by the rule. A rule can be viewed as placing mutual constraints on the tuples of objects that an action type can be applied to. Note that a single rule may allow no actions or many actions of one type. Given a decision list of such rules we say that an action is allowed by the list if it is allowed by some rule in the list, and no previous rule allows any actions. Again, a decision list may allow no actions or multiple actions of one type. A decision list $L$ for an MDP defines a deterministic policy $\pi[L]$ for that MDP. If $L$ allows no actions in state $s$, then $\pi[L](s)$ is the least[7] *legal* action in $s$; otherwise, $\pi[L](s)$ is the least legal action that is allowed by $L$. It is important to note that since $\pi[L]$ only considers legal actions, as specified by action preconditions, the rules do not need to encode the preconditions, which allows for simpler rules and learning. In other words, we can think of each rule as implicitly containing the preconditions of its action type.

As an example of a taxonomic decision list policy consider a simple blocks-world domain where the goal condition is always to clear off all of the red blocks. The primitive classes in this domain are **red**, **clear**, and **holding**, and the single relation is **on**. The following policy will solve any problem in the domain.

$$
\begin{aligned}
\textbf{putdown}(x_1) \quad &: \quad x_1 \in \textbf{holding} \\
\textbf{pickup}(x_1) \quad &: \quad x_1 \in \textbf{clear}, x_1 \in (\textbf{on}^*(\textbf{on red}))
\end{aligned}
$$

---

7. The action ordering in a relational MDP is defined lexicographically in terms of orderings on the action types and objects.

The first rule will cause the agent to putdown any block that is being held. Otherwise, if no block is being held, then find a block $x_1$ that is clear and is above a red block (expressed by **on\***(**on red**))) and pick it up. Appendix B gives examples of more complex policies that are learned by our system in the experiments.

### 5.3 Learning Taxonomic Decision Lists

For a given relational MDP, define $\mathcal{R}_{d,l}$ to be the set of action-selection rules that have a length of at most $l$ literals and whose class expression have depth at most $d$. Also, let $H_{d,l}$ denote the policy space defined by decision lists whose rules are from $\mathcal{R}_{d,l}$. Since the number of depth-bounded class expressions is finite there are a finite number of rules, and hence $H_{d,l}$ is finite, though exponentially large. Our implementation of **Learn-Policy**, as used in the main API loop, learns a policy in $H_{d,l}$ for user specified values of $d$ and $l$.

We use a Rivest-style decision-list learning approach (Rivest, 1987)—an approach also taken by Martin and Geffner (2000) for learning class-based policies. The primary difference between Martin and Geffner (2000) and our technique is the method for selecting individual rules in the decision list. We use a greedy, heuristic search, while previous work used an exhaustive enumeration approach. This difference allows us to find rules that are more complex, at the potential cost of failing to find some good simple rules that enumeration might discover.

Recall from Section 3, that the training set given to **Learn-Policy** contains trajectories of the rollout policy. Our learning algorithm, however, is not sensitive to the trajectory structure (i.e., the order of trajectory elements) and thus, to simplify our discussion, we will take the input to our learner to be a training set $D$ that contains the union of all the trajectory elements. This means that for a trajectory set that contains $n$ length $h$ trajectories, $D$ will contain a total of $n \cdot h$ training examples. As described in Section 3, each training example in $D$ has the form $\langle s, \pi(s), \hat{Q}(s, a_1), \ldots, \hat{Q}(s, a_m)\rangle$, where $s$ is a state, $\pi(s)$ is the action selected in $s$ by the previous policy, and $\hat{Q}(s, a_i)$ is the Q-value estimate of $Q^{\pi}(s, a_i)$. Note that in our experiments the training examples only contain values for the legal actions in a state.

Given a training set $D$, a natural learning goal is to find a decision-list policy that for each training example selects an action with the maximum estimated Q-value. This learning goal, however, can be problematic in practice as often there are several best (or close to best) actions as measured by the true Q-function. In such case, due to random sampling, the particular action that looks best according to the Q-value estimates in the training set is arbitrary. Attempting to learn a concise policy that matches these arbitrary actions will be difficult at best and likely impossible.

One approach (Lagoudakis & Parr, 2003) to avoiding this problem is to use statistical tests to determine the actions that are "clearly the best" (positive examples) and the ones that are "clearly not the best" (negative examples). The learner is then asked to find a policy that is consistent with the positive and negative examples. While this approach has shown some empirical success, it has the potential shortcoming of throwing away most of the Q-value information. In particular, it may not always be possible to find a policy that exactly matches the training data. In such cases, we would like the learner to make informed trade-offs regarding sub-optimal actions—i.e., prefer sub-optimal actions that have larger

---

**Learn-Decision-List** $(D, d, l, b)$

*// training set $D$, concept depth $d$, rule length $l$, beam width $b$*

$L \leftarrow$ **nil**;

**while** ($D$ is not empty)

    $R \leftarrow$ **Learn-Rule**$(D, d, l, b)$;

    $D \leftarrow D - \{d \in d \mid R \text{ covers } d\}$;

    $L \leftarrow$ **Extend-List**$(L, R)$; *// add $R$ to end of list*

**Return** $L$;

---

**Learn-Rule**$(D, d, l, b)$

*// training set $D$, concept depth $d$, rule length $l$, beam width $b$*

**for** each action type $a$    *// compute rule for each action type $a$*

    $R_a \leftarrow$ **Beam-Search**$(D, d, l, b, a)$;

**Return** $\text{argmax}_a$**Hvalue**$(R_a, D)$;

---

**Beam-Search** $(D, d, l, b, a)$

*// training set $D$, concept depth $d$, rule length $l$, beam width $b$, action type $a$*

$k \leftarrow$ arity of $a$; $X \leftarrow (x_1, \ldots, x_k)$;    *// $X$ is a sequence of action-argument variables*

$\text{L} \leftarrow \{(x \in C) \mid x \in X, C \in \mathcal{C}_d[X]\}$;    *// construct the set of depth bounded candidate literals*

$B_0 \leftarrow \{ a(X) : \textbf{nil} \}$; $i \leftarrow 1$;    *// initialize beam to a single rule with no literals*

**loop**

    $G = B_{i-1} \cup \{R \in \mathcal{R}_{d,l} \mid R = \textbf{Add-Literal}(R', l), R' \in B_{i-1}, l \in L\}$;

    $B_i \leftarrow$ **Beam-Select**$(G, b, D)$; *//   select best $b$ heuristic values*

    $i \leftarrow i + 1$;

**until** $B_{i-1} = B_i$; *//   loop until there is no more improvement in heuristic*

**Return** $\text{argmax}_{R \in B_i}$**Hvalue**$(R, D)$   *//   return best rule in final beam*

---

Figure 2: Pseudo-code for learning a decision list in $H_{d,l}$ given training data $D$. The procedure **Add-Literal**$(R, l)$ simply returns a rule where literal $l$ is added to the end of rule $R$. The procedure **Beam-Select**$(G, w, D)$ selects the best $b$ rules in $G$ with different heuristic values. The procedure **Hvalue**$(R, D)$ returns the heuristic value of rule $R$ relative to training data $D$ and is described in the text.

Q-values. With this motivation, below we describe a cost-sensitive decision-list learner that is sensitive to the full set of Q-values in $D$. The learning goal is roughly to find a decision list that selects actions with large cumulative Q-value over the training set.

**Learning List of Rules.** We say that a decision list $L$ covers a training example $\langle s, \pi(s), \hat{Q}(s, a_1), \ldots, \hat{Q}(s, a_m) \rangle$ if $L$ suggests an action in state $s$. Given a set of training examples $D$, we search for a decision list that selects actions with high Q-value via an iterative set-covering approach carried out by **Learn-Decision-List**. Decision-list rules are constructed one at a time and in order until the list covers all of the training examples. Pseudo-code for our algorithm is given in Figure 2. Initially, the decision list is the null list and does not cover any training examples. During each iteration, we search for a high quality rule $R$ with quality measured relative to the set of currently uncovered training examples. The selected rule is appended to the current decision-list, and the training examples newly covered by the selected rule are removed from the training set. This process repeats until the list covers all of the training examples. The success of this approach depends heavily on the function **Learn-Rule**, which selects a good rule relative to the uncovered training examples—typically a good rule is one that selects actions with the best (or close to best) Q-value and also covers a significant number of examples.

**Learning Individual Rules.** The input to the rule learner **Learn-Rule** is a set of training examples, along with depth and length parameters $d$ and $l$, and a beam width $b$. For each action type $a$, the rule learner calls the routine **Beam-Search** to find a good rule $R_a$ in $\mathcal{R}_{d,l}$ for action type $a$. **Learn-Rule** then returns the rule $R_a$ with the highest value as measured by our heuristic, which is described later in this section.

For a given action type $a$, the procedure **Beam-Search** generates a beam $B_0, B_1 \ldots$, where each $B_i$ is a set of rules in $\mathcal{R}_{d,l}$ for action type $a$. The sets evolve by specializing rules in previous sets by adding literals to them, guided by our heuristic function. Search begins with the most general rule $a(X) : \mathbf{nil}$, which allows any action of type $a$ in any state. Search iteration $i$ produces a set $B_i$ that contains $b$ rules with the highest *different* heuristic values among those in the following set[8]

$$G = B_{i-1} \cup \{R \in \mathcal{R}_{d,l} \mid R = \mathbf{Add\text{-}Literal}(R', l), R' \in B_{i-1}, l \in L\}$$

where $L$ is the set of all possible literals with a depth of $d$ or less. This set includes the current best rules (those in $B_{i-1}$) and also any rule in $\mathcal{R}_{d,l}$ that can be formed by adding a new literal to a rule in $B_{i-1}$. The search ends when no improvement in heuristic value occurs, that is when $B_i = B_{i-1}$. **Beam-Search** then returns the best rule in $B_i$ according to the heuristic.

**Heuristic Function.** For a training instance $\langle s, \pi(s), \hat{Q}(s, a_1), \ldots, \hat{Q}(s, a_m) \rangle$, we define the *Q-advantage* of taking action $a_i$ instead of $\pi(s)$ in state $s$ by $\Delta(s, a_i) = \hat{Q}(s, a_i) - \hat{Q}(s, \pi(s))$. Likewise, the Q-advantage of a rule $R$ is the sum of the Q-advantages of actions allowed by $R$ in $s$. Given a rule $R$ and a set of training examples $D$, our heuristic function **Hvalue**$(R, D)$ is equal to the number of training examples that the rule covers plus the cumulative Q-advantage of the rule over the training examples.[9] Using Q-advantage rather than Q-value focuses the learner toward instances where large improvement over the previous policy is possible. Naturally, one could consider using different weights for the coverage and Q-advantage terms, possibly tuning the weight automatically using validation data.

---

8. Since many rules in $\mathcal{R}_{d,l}$ are equivalent, we must prevent the beam from filling up with semantically equivalent rules. Rather than deal with this problem via expensive equivalence testing we take an ad-hoc, but practically effective approach. We assume that rules do not coincidentally have the same heuristic value, so that ones that do must be equivalent. Thus, we construct beams whose members all have different heuristic values. We choose between rules with the same value by preferring shorter rules, then arbitrarily.

9. If the coverage term is not included, then covering a zero Q-advantage example is the same as not covering it. But zero Q-advantage can be good (e.g., the previous policy is optimal in that state).

## 6. Random Walk Bootstrapping

There are two issues that are critical to the success of our API technique. First, API is fundamentally limited by the expressiveness of the policy language and the strength of the learner, which dictates its ability to capture the improved policy described by the training data at each iteration. Second, API can only yield improvement if **Improved-Trajectories** successfully generates training data that describes an improved policy. For large classical planning domains, initializing API with an uninformed random policy will typically result in essentially random training data, which is not helpful for policy improvement. For example, consider the MDP corresponding to the 20-block blocks world with an initial problem distribution that generates random initial and goal states. In this case, a random policy is unlikely to reach a goal state within any practical horizon time. Hence, the rollout trajectories are unlikely to reach the goal, providing no guidance toward learning an improved policy (i.e., a policy that can more reliably reach the goal).

Because we are interested in solving large domains such as this, providing guiding inputs to API is critical. In Fern, Yoon, and Givan (2003), we showed that by bootstrapping API with the domain-independent heuristic of the planner FF (Hoffmann & Nebel, 2001), API was able to uncover good policies for the blocks world, simplified logistics world (no planes), and stochastic variants. This approach, however, is limited by the heuristic's ability to provide useful guidance, which can vary widely across domains.

Here we describe a new bootstrapping procedure for goal-based planning domains, based on random walks, for guiding API toward good policies. Our planning system, which is evaluated in Section 7, is based on integrating this procedure with API in order to find policies for goal-based planning domains. For non-goal-based MDPs, this bootstrapping procedure can not be directly applied, and other bootstrapping mechanisms must be used if necessary. This might include providing an initial non-trivial policy, providing a heuristic function, or some form of reward shaping (Mataric, 1994). Below, we first describe the idea of random-walk distributions. Next, we describe how to use these distributions in the context of bootstrapping API, giving a new algorithm **LRW-API**.

### 6.1 Random Walk Distributions

Throughout we consider an MDP $M = \langle S, A, T, R, I \rangle$ that correspond to goal-based planning domains, as described in Section 5.1. Recall that each state $s \in S$ corresponds to a planning problem, specifying a world state (via world facts) and a set of goal conditions (via goal facts). We will use the terms "MDP state" and "planning problem" interchangeably. Note that, in this context, $I$ is a distribution over planning problems. For convenience we will denote MDP states as tuples $s = (w, g)$, where $w$ and $g$ are the sets of world facts and goal facts in $s$ respectively.

Given an MDP state $s = (w, g)$ and set of goal predicates $G$, we define $s|_G$ to be the MDP state $(w, g')$ where $g'$ contains those goal facts in $g$ that are applications of a predicate in $G$. Given $M$ and a set of *goal predicates* $G$, we define the *n-step random-walk problem distribution* $\mathcal{RW}_n(M, G)$ by the following stochastic algorithm:

1. Draw a random state $s_0 = (w_0, g_0)$ from the initial state distribution $I$.

2. Starting at $s_0$ take $n$ uniformly random actions[10], giving a state sequence $(s_0, \ldots, s_n)$, where $s_n = (w_n, g_0)$ (recall that actions do not change goal facts). At each uniformly random action selection, we assume that an extra "no-op" action (that does not change the state) is selected with some fixed probability, for reasons explained below.

3. Let $g$ be the set of goal facts corresponding to the world facts in $w_n$, so e.g., if $w_n = \{\textbf{on}(a, b), \textbf{clear}(a)\}$, then $g = \{\textbf{gon}(a, b), \textbf{gclear}(a)\}$. Return the planning problem (MDP state) $(s_0, g)|_G$ as the output.

We will sometimes abbreviate $\mathcal{RW}_n(M, G)$ by $\mathcal{RW}_n$ when $M$ and $G$ are clear in context.

Intuitively, to perform well on this distribution a policy must be able to achieve facts involving the goal predicates that typically result after an $n$-step random walk from an initial state. By restricting the set of goal predicates $G$ we can specify the types of facts that we are interested in achieving—e.g., in the blocks world we may only be interested in achieving facts involving the "on" predicate.

The random-walk distributions provide a natural way to span a range of problem difficulties. Since longer random walks tend to take us further from an initial state, for small $n$ we typically expect that the planning problems generated by $\mathcal{RW}_n$ will become more difficult as $n$ grows. However, as $n$ becomes large, the problems generated will require far fewer than $n$ steps to solve—i.e., there will be more direct paths from an initial state to the end state of a long random walk. Eventually, since $S$ is finite, the problem difficulty will stop increasing with $n$.

A question raised by this idea is whether, for large $n$, good performance on $\mathcal{RW}_n$ ensures good performance on other problem distributions of interest in the domain. In some domains, such as the simple blocks world[11], good random-walk performance does seem to yield good performance on other distributions of interest. In other domains, such as the grid world (with keys and locked doors), intuitively, a random walk is very unlikely to uncover a problem that requires unlocking a sequence of doors. Indeed, since $\mathcal{RW}_n$ is insensitive to the goal distribution of the underlying planning domain, the random-walk distribution may be quite different.

We believe that good performance on long random walks is often useful, but is only addressing one component of the difficulty of many planning benchmarks. To successfully address problems with other components of difficulty, a planner will need to deploy orthogonal technology such as landmark extraction for setting subgoals (Hoffman, Porteous, & Sebastia, 2004). For example, in the grid world, if we could automatically set the subgoal of possessing a key for the first door, a long random-walk policy could provide a useful macro for getting that key.

For the purpose of developing a bootstrapping technique for API, we limit our focus to finding good policies for long random walks. In our experiments, we define "long" by specifying a large walk length $N$. Theoretically, the inclusion of the "no-op" action in the definition of $\mathcal{RW}$ ensures that the induced random-walk Markov chain[12] is aperiodic, and thus that the distribution over states reached by increasingly long random walks converges to a stationary distribution[13]. Thus $\mathcal{RW}_* = \lim_{n\to\infty} \mathcal{RW}_n$ is well-defined, and we take good performance on $\mathcal{RW}_*$ to be our goal.

---

10. In practice, we only select random actions from the set of applicable actions in a state $s_i$, provided our simulator makes it possible to identify this set.

11. In the blocks world with large $n$, $\mathcal{RW}_n$ generates various pairs of random block configurations, typically pairing states that are far apart—clearly, a policy that performs well on this distribution has captured significant information about the blocks world.

12. We don't formalize this chain here, but various formalizations work well.

## 6.2 Random-Walk Bootstrapping

For an MDP $M$, we define $M[I']$ to be an MDP identical to $M$ only with the initial state distribution replaced by $I'$. We also define the *success ratio* $\text{SR}(\pi, M[I])$ of $\pi$ on $M[I]$ as the probability that $\pi$ solves a problem drawn from $I$. Also treating $I$ as a random variable, the *average length* $\text{AL}(\pi, M[I])$ of $\pi$ on $M[I]$ is the conditional expectation of the solution length of $\pi$ on problems drawn from $I$ given that $\pi$ solves $I$. Typically the solution length of a problem is taken to be the number of actions, however, when action costs are not uniform, the length is taken to be the sum of the action costs. Note that for the MDP formulation of classical planning domains, given in Section 5.1, if a policy $\pi$ achieves a high $\overline{V}(\pi)$ then it will also have a high success ratio and low average cost.

Given an MDP $M$ and set of goal predicates $G$, our system attempts to find a good policy for $M[\mathcal{RW}_N]$, where $N$ is selected to be large enough to adequately approximate $\mathcal{RW}_*$, while still allowing tractable completion of the learning. Naively, given an initial random policy $\pi_0$, we could try to apply API directly. However, as already discussed, this will not work in general, since we are interested in planning domains where $\mathcal{RW}_*$ produces extremely large and difficult problems where random policies provide an ineffective starting point.

However, for very small $n$ (e.g., $n = 1$), $\mathcal{RW}_n$ typically generates easy problems, and it is likely that API, starting with even a random initial policy, can reliably find a good policy for $\mathcal{RW}_n$. Furthermore, we expect that if a policy $\pi_n$ performs well on $\mathcal{RW}_n$, then it will also provide reasonably good, but perhaps not perfect, guidance on problems drawn from $\mathcal{RW}_m$ when $m$ is only moderately larger than $n$. Thus, we expect to be able to find a good policy for $\mathcal{RW}_m$ by bootstrapping API with initial policy $\pi_n$. This suggests a natural iterative bootstrapping technique to find a good policy for large $n$ (in particular, for $n = N$).

Figure 3 gives pseudo-code for the procedure **LRW-API** which integrates API and random-walk bootstrapping to find a policy for the long-random-walk problem distribution. Intuitively, this algorithm can be viewed as iterating through two stages: first, finding a hard enough distribution for the current policy (by increasing $n$); and, then, finding a good policy for the hard distribution using API. The algorithm maintains a current policy $\pi$ and current walk length $n$ (initially $n = 1$). As long as the success ratio of $\pi$ on $RW_n$ is below the *success threshold* $\tau$, which is a constant close to one, we simply iterate steps of approximate policy improvement. Once we achieve a success ratio of $\tau$ with some policy $\pi$, the if-statement increases $n$ until the success ratio of $\pi$ on $\mathcal{RW}_n$ falls below $\tau - \delta$. That is, when $\pi$ performs well enough on the current $n$-step distribution we move on to a distribution that is slightly harder. The constant $\delta$ determines how much harder and is set small enough so that $\pi$ can likely be used to bootstrap policy improvement on the harder distribution. (The simpler method of just increasing $n$ by 1 whenever success ratio $\tau$ is achieved will also

---

13. The Markov chain may not be irreducible, so the same stationary distribution may not be reached from all initial states; however, we are only considering one initial state, described by $I$.

```
LRW-API (N, G, n, w, h, M, π₀, γ)

//  max random-walk length N, goal predicates G
//  training set size n, sampling width w, horizon h,
//  MDP M, initial policy π₀, discount factor γ.

π ← π₀;  n ← 1;

loop
    if  ŜR_π(n) > τ

        // Find harder n-step distribution for π.
        n ← least i ∈ [n, N] s.t. ŜR_π(i) < τ − δ, or N if none;

    M' = M[RWₙ(M, G)];
    T ← Improved-Trajectories(n, w, h, M', π);
    π ← Learn-Policy(T);

until  satisfied with π

Return π;
```

Figure 3: Pseudo-code for **LRW-API**. $\widehat{\mathrm{SR}}_\pi(n)$ estimates the success ratio of $\pi$ in planning domain $D$ on problems drawn from $\mathcal{RW}_n(M, G)$ by drawing a set of problems and returning the fraction solved by $\pi$. Constants $\tau$ and $\delta$ are described in the text.

find good policies whenever this method does. This can take much longer, as it may run API repeatedly on a training set for which we already have a good policy.)

Once $n$ becomes equal to the maximum walk length $N$, we will have $n = N$ for all future iterations. It is important to note that even after we find a policy with a good success ratio on $\mathcal{RW}_N$ it may still be possible to improve on the average length of the policy. Thus, we continue API on this distribution until we are satisfied with both the success ratio and average length of the current policy.

## 7. Relational Planning Experiments

In this section, we evaluate the **LRW-API** technique on relational MDPs corresponding to deterministic and stochastic classical planning domains. We first give results for a number of deterministic benchmark domains, showing promising results in comparison with the state-of-the-art planner FF (Hoffmann & Nebel, 2001), while also highlighting limitations of our approach. Next, we give results for several stochastic planning domains including those in the domain-specific track of the 2004 International Probabilistic Planning Competition (IPPC). All of the domain definitions and problem generators used in our experiments are available upon request.

In all of our experiments, we use the policy learner described in Section 5.3 to learn taxonomic decision list policies. In all cases, the number of training trajectories is 100, and policies are restricted to rules with a depth bound $d$ and length bound $l$. The discount factor $\gamma$ was always one, and **LRW-API** was always initialized with a policy that selects random actions. We utilize a maximum-walk-length parameter $N = 10,000$ and set $\tau$ and $\delta$ equal to 0.9 and 0.1 respectively.

## 7.1 Deterministic Planning Experiments

We perform experiments in seven familiar STRIPS planning domains including those used in the AIPS-2000 planning competition, those used to evaluate TL-Plan in Bacchus and Kabanza (2000), and the Gripper domain. Each domain has a standard problem generator that accepts parameters, which control the size and difficulty of the randomly generated problems. Below we list each domain and the parameters associated with them. A detailed description of these domains can be found in Hoffmann and Nebel (2001).

- Blocks World $(n)$ : the standard blocks worlds with $n$ blocks.

- Freecell $(s, c, f, l)$ : a version of Solitaire with $s$ suits, $c$ cards per suit, $f$ freecells, and $l$ columns.

- Logistics (a,c,l,p) : the logistics transportation domain with $a$ airplanes, $c$ cities, $l$ locations, and $p$ packages.

- Schedule $(p)$ : a job shop scheduling domain with $p$ parts.

- Elevator $(f, p)$ : elevator scheduling with $f$ floors and $p$ people.

- Gripper $(b)$ : a robotic gripper domain with $b$ balls.

- Briefcase $(i)$ : a transportation domain with $i$ items.

**LRW Experiments.** Our first set of experiments evaluates the ability of **LRW-API** to find good policies for $\mathcal{RW}_*$. Here we utilize a sampling width of one for rollout, since these are deterministic domains. Recall that in each iteration of **LRW-API** we compute an (approximately) improved policy and may also increase the walk length $n$ to find a harder problem distribution. We continued iterating **LRW-API** until we observed no further improvement. The training time per iteration is approximately five hours.[14] Though the initial training period is significant, once a policy is learned it can be used to solve new problems very quickly, terminating in seconds with a solution when one is found, even for very large problems.

Figure 4 provides data for each iteration of **LRW-API** in each of the seven domains with the indicated parameter settings. The first column, for each domain, indicates the iteration number (e.g., the Blocks World was run for 8 iterations). The second column records the walk length $n$ used for learning in the corresponding iteration. The third and fourth columns record the SR and AL of the policy learned at the corresponding iteration as measured on 100 problems drawn from $\mathcal{RW}_n$ for the corresponding value of $n$ (i.e., the distribution used for learning). When this SR exceeds $\tau$, the next iteration seeks an increased walk length $n$. The fifth and sixth columns record the SR and AL of the same policy, but measured on 100 problems drawn from the LRW target distribution $\mathcal{RW}_*$, which in these experiments is approximated by $\mathcal{RW}_N$ for $N = 10,000$.

---

14. This timing information is for a relatively unoptimized Scheme implementation. A reimplementation in C would likely result in a 5-10 fold speed-up.

| iter. # | $n$ | $\mathcal{RW}_n$ SR | $\mathcal{RW}_n$ AL | $\mathcal{RW}_*$ SR | $\mathcal{RW}_*$ AL |
|---|---|---|---|---|---|
| **Blocks World (20)** | | | | | |
| 1 | 4 | 0.92 | 2.0 | 0 | 0 |
| 2 | 14 | 0.94 | 5.6 | 0.10 | 41.4 |
| 3 | 54 | 0.56 | 15.0 | 0.17 | 42.8 |
| 4 | 54 | 0.78 | 15.0 | 0.32 | 40.2 |
| 5 | 54 | 0.88 | 33.7 | 0.65 | 47.0 |
| 6 | 54 | 0.98 | 25.1 | 0.90 | 43.9 |
| 7 | 334 | 0.84 | 45.6 | 0.87 | 50.1 |
| 8 | 334 | 0.99 | 37.8 | 1 | 43.3 |
| FF | | | | 0.96 | 49.0 |
| **Freecell (4,2,2,4)** | | | | | |
| 1 | 5 | 0.97 | 1.4 | 0.08 | 3.6 |
| 2 | 8 | 0.97 | 2.7 | 0.26 | 6.3 |
| 3 | 30 | 0.65 | 7.0 | 0.78 | 7.0 |
| 4 | 30 | 0.72 | 7.1 | 0.85 | 7.0 |
| 5 | 30 | 0.90 | 6.7 | 0.85 | 6.3 |
| 6 | 30 | 0.81 | 6.7 | 0.89 | 6.6 |
| 7 | 30 | 0.78 | 6.8 | 0.87 | 6.8 |
| 8 | 30 | 0.90 | 6.9 | 0.89 | 6.6 |
| 9 | 30 | 0.93 | 7.7 | 0.93 | 7.9 |
| FF | | | | 1 | 5.4 |
| **Elevator (20,10)** | | | | | |
| 1 | 20 | 1 | 4.0 | 1 | 26 |
| FF | | | | 1 | 23 |
| **Gripper (10)** | | | | | |
| 1 | 10 | 1 | 3.8 | 1 | 13 |
| FF | | | | 1 | 13 |

| iter. # | $n$ | $\mathcal{RW}_n$ SR | $\mathcal{RW}_n$ AL | $\mathcal{RW}_*$ SR | $\mathcal{RW}_*$ AL |
|---|---|---|---|---|---|
| **Logistics (1,2,2,6)** | | | | | |
| 1 | 5 | 0.86 | 3.1 | 0.25 | 11.3 |
| 2 | 45 | 0.86 | 6.5 | 0.28 | 7.2 |
| 3 | 45 | 0.81 | 6.9 | 0.31 | 8.4 |
| 4 | 45 | 0.86 | 6.8 | 0.28 | 8.9 |
| 5 | 45 | 0.76 | 6.1 | 0.28 | 7.8 |
| 6 | 45 | 0.76 | 5.9 | 0.32 | 8.4 |
| 7 | 45 | 0.86 | 6.2 | 0.39 | 9.1 |
| 8 | 45 | 0.76 | 6.9 | 0.31 | 11.0 |
| 9 | 45 | 0.70 | 6.1 | 0.19 | 7.8 |
| 10 | 45 | 0.81 | 6.1 | 0.25 | 7.6 |
| … | … | … | … | … | … |
| 43 | 45 | 0.74 | 6.4 | 0.25 | 9.0 |
| 44 | 45 | 0.90 | 6.9 | 0.39 | 9.3 |
| 45 | 45 | 0.92 | 6.6 | 0.38 | 9.4 |
| FF | | | | 1 | 13 |
| **Schedule (20)** | | | | | |
| 1 | 1 | 0.79 | 1 | 0.48 | 27 |
| 2 | 4 | 1 | 3.45 | 1 | 34 |
| FF | | | | 1 | 36 |
| **Briefcase (10)** | | | | | |
| 1 | 5 | 0.91 | 1.4 | 0 | 0 |
| 2 | 15 | 0.89 | 4.2 | 0.2 | 38 |
| 3 | 15 | 1 | 3.0 | 1 | 30 |
| FF | | | | 1 | 28 |

Figure 4: Results for each iteration of **LRW-API** in seven deterministic planning domains. For each iteration, we show the walk length $n$ used for learning, along with the success ratio (SR) and average length (AL) of the learned policy on both $\mathcal{RW}_n$ and $\mathcal{RW}_*$. The final policy shown in each domain performs above $\tau = 0.9$ SR on walks of length $N = 10,000$ (with the exception of Logistics), and further iteration does not improve the performance. For each benchmark we also show the SR and AL of the planner FF on problems drawn from $\mathcal{RW}_*$.

So, for example, we see that in the Blocks World there are a total of 8 iterations, where we learn at first for one iteration with $n = 4$, one more iteration with $n = 14$, four iterations with $n = 54$, and then two iterations with $n = 334$. At this point we see that the resulting policy performs well on $\mathcal{RW}_*$. Further iterations with $n = N$, not shown, showed no improvement over the policy found after iteration eight. In other domains, we also observed no improvement after iterating with $n = N$, and thus do not show those iterations. We note that all domains except Logistics (see below) achieve policies with good performance on $\mathcal{RW}_N$ by learning on much shorter $\mathcal{RW}_n$ distributions, indicating that we have indeed selected a large enough value of $N$ to capture $\mathcal{RW}_*$, as desired.

**General Observations.** For several domains, our learner bootstraps very quickly from short random-walk problems, finding a policy that works well even for much longer random-walk problems. These include Schedule, Briefcase, Gripper, and Elevator. Typically, large problems in these domains have many somewhat independent subproblems with short solutions, so that short random walks can generate instances of all the different typical subproblems. In each of these domains, our best LRW policy is found in a small number of iterations and performs comparably to FF on $\mathcal{RW}_*$. We note that FF is considered a very good domain-independent planner for these domains, so we consider this a successful result.

For two domains, Logistics[15] and Freecell, our planner is unable to find a policy with success ratio one on $\mathcal{RW}_*$. We believe that this is a result of the limited knowledge representation we allowed for policies for the following reasons. First, we ourselves cannot write good policies for these domains within our current policy language. For example, in logistics, one of the important concept is "the set containing all packages on trucks such that the truck is in the packages goal city". However, the domain is defined in such a way that this concept cannot be expressed within the language used in our experiments. Second, the final learned decision lists for Logistics and Freecell, which are in Appendix B, contain a much larger number of more specific rules than the lists learned in the other domains. This indicates that the learner has difficulty finding general rules, within the language restrictions, that are applicable to large portions of training data, resulting in poor generalization. Third, the success ratio (not shown) for the sampling-based rollout policy, i.e., the improved policy simulated by **Improved-Trajectories**, is substantially higher than that for the resulting learned policy that becomes the policy of the next iteration. This indicates that **Learn-Decision-List** is learning a much weaker policy than the sampling-based policy generating its training data, indicating a weakness in either the policy language or the learning algorithm. For example, in the logistics domain, at iteration eight, the training data for learning the iteration-nine policy is generated by a sampling rollout policy that achieves success ratio 0.97 on 100 training problems drawn from the same $\mathcal{RW}_{45}$ distribution, but the learned iteration-nine policy only achieves success ratio 0.70, as shown in the figure at iteration nine. Extending our policy language to incorporate the expressiveness that appears to be required in these domains will require a more sophisticated learning algorithm, which is a point of future work.

---

15. In Logistics, the planner generates a long sequence of policies with similar, oscillating success ratio that are elided from the table with an ellipsis for space reasons.

| Domain | Size | $\pi_*$ SR | $\pi_*$ AL | FF SR | FF AL |
|--------|------|:---:|:---:|:---:|:---:|
| Blocks | (20) | 1 | 54 | 0.81 | 60 |
|  | (50) | 1 | 151 | 0.28 | 158 |
| Freecell | (4,2,2,4) | 0.36 | 15 | 1 | 10 |
|  | (4,13,4,8) | 0 | — | 0.47 | 112 |
| Logistics | (1,2,2,6) | 0.87 | 6 | 1 | 6 |
|  | (3,10,2,30) | 0 | — | 1 | 158 |
| Elevator | (60,30) | 1 | 112 | 1 | 98 |
| Schedule | (50) | 1 | 175 | 1 | 212 |
| Briefcase | (10) | 1 | 30 | 1 | 29 |
|  | (50) | 1 | 162 | 0 | — |
| Gripper | (50) | 1 | 149 | 1 | 149 |

Figure 5: Results on standard problem distributions for seven benchmarks. Success ratio (SR) and average length (AL) are provided for both FF and our policy learned for the LRW problem distribution. For a given domain, the same learned LRW policy is used for each problem size shown.

In the remaining domain, the Blocks World, the bootstrapping provided by increasingly long random walks appears particularly useful. The policies learned at each of the walk lengths 4, 14, 54, and 334 are increasingly effective on the target LRW distribution $\mathcal{RW}_*$. For walks of length 54 and 334, it takes multiple iterations to master the provided level of difficulty beyond the previous walk length. Finally, upon mastering walk length 334, the resulting policy appears to perform well for any walk length. The learned policy is modestly superior to FF on $\mathcal{RW}_*$ in success ratio and average length.

**Evaluation on the Original Problem Distributions.** In each domain we denote by $\pi_*$ the best learned LRW policy—i.e., the policy, from each domain, with the highest performance on $\mathcal{RW}_*$, as shown in Figure 4. The taxonomic decision lists corresponding to $\pi_*$ for each domain is given in Appendix B. Figure 5 shows the performance of $\pi_*$, in comparison to FF, on the original intended problem distributions for each of our domains. We measured the success ratio of both systems by giving a time limit of 100 seconds to solve a problem. Here we have attempted to select the largest problem sizes previously used in evaluation of domain-specific planners, either in AIPS-2000 or in Bacchus and Kabanza (2000), as well as show a smaller problem size for those cases where one of the planners we show performed poorly on the large size. In each case, we use the problem generators provided with the domains, and evaluate on 100 problems of each size.

Overall, these results indicate that our learned, reactive policies are competitive with the domain-independent planner FF. It is important to remember that these policies are learned in a domain-independent fashion, and thus **LRW-API** can be viewed as a general approach to generating domain-specific reactive planners. On two domains, Blocks World and Briefcase, our learned policies substantially outperform FF on success ratio, especially on large domain sizes. On three domains, Elevator, Schedule, and Gripper, the two approaches perform quite similarly on success ratio, with our approach superior in average length on Schedule but FF superior in average length on Elevator.

On two domains, Logistics and Freecell, FF substantially outperforms our learned policies on success ratio. We believe that this is partly due to an inadequate policy language, as discussed above. We also believe, however, that another reason for the poor performance is that the long-random-walk distribution $\mathcal{RW}_*$ does not correspond well to the standard problem distributions. This seems to be particularly true for Freecell. The policy learned for Freecell (4,2,2,4) achieved a success ratio of 93 percent on $\mathcal{RW}_*$, however, for the standard distribution it only achieved 36 percent. This suggests that $\mathcal{RW}_*$ generates problems that are significantly easier than the standard distribution. This is supported by the fact that the solutions produced by FF on the standard distribution are on average twice as long as those produced on $\mathcal{RW}_*$. One likely reason for this is that it is easy for random walks to end up in dead states in Freecell, where no actions are applicable. Thus the random walk distribution will typically produce many problems where the goals correspond to such dead states. The standard distribution on the other hand will not treat such dead states as goals.

## 7.2 Probabilistic Planning Experiments

Here we present experiments in three probabilistic domains that are described in the probabilistic planning domain language PPDDL (Younes, 2003).

- Ground Logistics $(c, p)$ : a probabilistic version of logistics with no airplanes, with $c$ cities and $p$ packages. The driving action has a probability of failure in this domain.

- Colored Blocks World $(n)$ : a probabilistic blocks world with $n$ colored blocks, where goals involve constructing towers with certain color patterns. There is a probability that moved blocks fall to the floor.

- Boxworld $(c, p)$ : a probabilistic version of full logistics with $c$ cities and $p$ packages. Transportation actions have a probability of going in the wrong direction.

The Ground Logistics domain is originally from Boutilier et al. (2001), and was also used for evaluation in Yoon et al. (2002). The Colored Blocks World and Boxworld domains are the domains used in the hand-tailored track of IPPC in which our LRW-API technique was entered. In the hand-tailored track, participants were provided with problem generators for each domain before the competition and were allowed to incorporate domain knowledge into the planner for use at competition time. We provided the problem generators to LRW-API and learned policies for these domains, which were then entered into the competition.

We have also conducted experiments in the other probabilistic domains from Yoon et al. (2002), including variants of the blocks world and a variant of Ground Logistics, some of which appeared in Fern et al. (2003). However, we do not show those results here since they are qualitatively identical to the deterministic blocks world results described above and the Ground Logistics results we show below.

For our three probabilistic domains, we conducted LRW experiments using the same procedure as above. All parameters given to **LRW-API** were the same as above except that the sampling width used for rollout was set to $w = 10$, and $\tau$ was set to $0.85$ in order to account for the stochasticity in these domains. The results of these experiments are shown in Figure 6. These tables have the same form as Figure 4 only the last row given for each domain now gives the performance of $\pi_*$ on the standard distribution, i.e., problems drawn from the domains problem generator. For Colored Blocks World the problem generator produces problems whose goals are specified using existential quantifiers. For example, a simple goal may be "there exists blocks $x$ and $y$ such that $x$ is red, $y$ is blue and $x$ is on $y$". Since our policy language cannot directly handle existentially quantified goals we preprocess the planning problems produced by the problem generator to remove them. This was done by assigning particular block names to the existential variables, ensuring that the static properties of a block (in this case color) satisfied the static properties of the variable is was assigned to. In this domain, finding such an assignment was trivial, and the resulting assignment was taken to be the goal, giving a planning problem to which our learned policy was applied. Since the blocks world states are fully connected, the resulting goal is always guaranteed to be achievable.

| iter. # | $n$ | $\mathcal{RW}_n$ | | $\mathcal{RW}_*$ | |
|---|---|---|---|---|---|
| | | SR | AL | SR | AL |
| | | **Boxworld (10,5)** | | | |
| 1 | 10 | 0.73 | 4.3 | 0.03 | 61.5 |
| 2 | 10 | 0.93 | 2.3 | 0.13 | 58.4 |
| 3 | 20 | 0.91 | 4.4 | 0.17 | 55.9 |
| 4 | 40 | 0.96 | 6.1 | 0.31 | 50.4 |
| 5 | 170 | 0.62 | 30.8 | 0.25 | 52.2 |
| 6 | 170 | 0.49 | 37.9 | 0.17 | 55.7 |
| 7 | 170 | 0.63 | 29.3 | 0.21 | 55 |
| 8 | 170 | 0.63 | 29.1 | 0.18 | 55.3 |
| 9 | 170 | 0.48 | 36.4 | 0.17 | 55.3 |
| Standard Distribution (15,15) | | | | 0 | – |

| iter. # | $n$ | $\mathcal{RW}_n$ | | $\mathcal{RW}_*$ | |
|---|---|---|---|---|---|
| | | SR | AL | SR | AL |
| | | **Ground Logistics (3,4,4,3)** | | | |
| 1 | 5 | 0.95 | 2.71 | 0.17 | 168.9 |
| 2 | 10 | 0.97 | 2.06 | 0.84 | 17.5 |
| 3 | 160 | 1 | 6.41 | 1 | 7.2 |
| Standard Distribution (5,7,7,20) | | | | 1 | 20 |
| | | **Colored Blocks World (10)** | | | |
| 1 | 2 | 0.86 | 1.7 | 0.19 | 93.6 |
| 2 | 5 | 0.89 | 8.4 | 0.81 | 40.8 |
| 3 | 40 | 0.92 | 11.7 | 0.85 | 32.7 |
| 4 | 100 | 0.76 | 37.5 | 0.77 | 38.5 |
| 5 | 100 | 0.94 | 20.0 | 0.95 | 21.9 |
| Standard Distribution (50) | | | | 0.95 | 123 |

Figure 6: Results for each iteration of **LRW-API** in three probabilistic planning domains. For each iteration, we show the walk length $n$ used for learning, along with the success ratio (SR) and average length (AL) of the learned policy on both $\mathcal{RW}_n$ and $\mathcal{RW}_*$. For each benchmark, we show performance on the standard problem distribution of the policy whose performance is best on $\mathcal{RW}_*$.

For Boxworld, **LRW-API** is not able to find a good policy for $\mathcal{RW}_*$ or the standard distribution. Again, as for deterministic Logistics and Freecell, we believe that this is primarily because of the restricted policy languages that is currently used by our learner. Here, as for those domains, we see that the decision list learned for Boxworld contains many very specific rules, indicating that the learner was not able to generalize well beyond the training trajectories. For Ground Logistics, we see that **LRW-API** quickly finds a good policy for both $\mathcal{RW}_*$ and the standard distribution.

For Colored Blocks World, we also see that **LRW-API** is able to quickly find a good policy for both $\mathcal{RW}_*$ and the standard distribution. However, unlike the deterministic (uncolored) blocks world, here the success ratio is observed to be less than one, solving 95 percent of the problems. It is unclear, why **LRW-API** is not able to find a "perfect" policy. It is relatively easy to hand-code a policy for Colored Blocks World using the language of the learner, hence inadequate knowledge representation is not the answer. The predicates and action types for this domain are not the same as those in its deterministic counterpart and other stochastic variants that we have previously considered. This difference apparently interacts badly with our learners search bias, causing it to fail to find a perfect policy. Nevertheless, these two results, along with the probabilistic planning results not shown here, indicate that when a good policy is expressible in our language, **LRW-API** can find good policies in complex relational MDPs. This makes **LRW-API** one of the few techniques that can simultaneously cope with the complexity resulting from stochasticity and from relational structure in domains such as these.

## 8. Related Work

Boutilier et al. (2001) presented the first exact solution technique for relational MDPs based on structured dynamic programming. However, a practical implementation of the approach was not provided, primarily due to the need for the simplification of first-order logic formulas. These ideas, however, served as the basis for a logic-programming-based system (Kersting, Van Otterlo, & DeRaedt, 2004) that was successfully applied to blocks-world problems involving simple goals and a simplified logistics world. This style of approach is inherently limited to domains where the exact value functions and/or policies can be compactly represented in the chosen knowledge representation. Unfortunately, this is not generally the case for the types of domains that we consider here, particularly as the planning horizon grows. Nevertheless, providing techniques such as these that directly reason about the MDP model is an important direction. Note that our API approach essentially ignores the underlying MDP model, and simply interacts with the MDP simulator as a black box.

An interesting research direction is to consider principled approximations of these techniques that can discover good policies in more difficult domains. This has been considered by Guestrin et al. (2003a), where a class-based MDP and value function representation was used to compute an approximate value function that could generalize across different sets of objects. Promising empirical results were shown in a multi-agent tactical battle domain. Presently the class-based representation does not support some of the representation features that are commonly found in classical planning domains (e.g., relational facts such as $\mathbf{on}(a, b)$ that change over time), and thus is not directly applicable in these contexts. However, extending this work to richer representations is an interesting direction. Its ability to "reason globally" about a domain may give it some advantages compared to API.

Our approach is closely related to work in relational reinforcement learning (RRL) (Dzeroski et al., 2001), a form of online API that learns relational value-function approximations. $Q$-value functions are learned in the form of relational decision trees ($Q$-trees) and are used to learn corresponding policies ($P$-trees). The RRL results clearly demonstrate the difficulty of learning value-function approximations in relational domains. Compared to $P$-trees, $Q$-trees tend to generalize poorly and be much larger. RRL has not yet demonstrated scalability to problems as complex as those considered here—previous RRL blocks-world experiments include relatively simple goals[16], which lead to value functions that are much less complex than the ones here. For this reason, we suspect that RRL would have difficulty in the domains we consider, precisely because of the value-function approximation step that we avoid; however, this needs to be experimentally tested.

We note, however, that our API approach has the advantage of using an unconstrained simulator, whereas RRL learns from irreversible world experience (pure RL). By using a simulator, we are able to estimate the $Q$-values for *all* actions at each training state, providing us with rich training data. Without such a simulator, RRL is not able to directly estimate the $Q$-value for each action in each training state—thus, RRL learns a $Q$-tree to provide estimates of the $Q$-value information needed to learn the $P$-tree. In this way, value-function learning serves a more critical role when a simulator is unavailable. We believe, that in many relational planning problems, it is possible to learn a model or simulator from world experience—in this case, our API approach can be incorporated as the planning component of RRL. Otherwise, finding ways to either avoid learning or to more effectively learn relational value-functions in RRL is an interesting research direction.

Researchers in classical planning have long studied techniques for learning to improve planning performance. For a collection and survey of work on "learning for planning domains" see Minton (1993) and Zimmerman and Kambhampati (2003). Two primary approaches are to learn domain-specific control rules for guiding search-based planners e.g., Minton, Carbonell, Knoblock, Kuokka, Etzioni, and Gil (1989), Veloso, Carbonell, Perez, Borrajo, Fink, and Blythe (1995), Estlin and Mooney (1996), Huang, Selman, and Kautz (2000), Ambite, Knoblock, and Minton (2000), Aler, Borrajo, and Isasi (2002), and, more closely related, to learn domain-specific reactive control policies (Khardon, 1999a; Martin & Geffner, 2000; Yoon et al., 2002).

Regarding the latter, our work is novel in using API to iteratively improve stand-alone control policies. Regarding the former, in theory, search-based planners can be iteratively improved by continually adding newly learned control knowledge—however, it can be difficult to avoid the utility problem (Minton, 1988), i.e., being swamped by low utility rules. Critically, our policy-language bias confronts this issue by preferring simpler policies. Our learning approach is also not tied to having a base planner (let alone tied to a single particular base planner), unlike most previous work. Rather, we only require a domain simulator.

The ultimate goal of such systems is to allow for planning in large, difficult problems that are beyond the reach of domain-independent planning technology. Clearly, learning to achieve this goal requires some form of bootstrapping and almost all previous systems have relied on the human for this purpose. By far, the most common human-bootstrapping approach is "learning from small problems". Here, the human provides a small problem distribution to the learner, by limiting the number of objects (e.g., using 2-5 blocks in the blocks world), and control knowledge is learned for the small problems. For this approach to work, the human must ensure that the small distribution is such that good control knowledge for the small problems is also good for the large target distribution. In contrast, our long-random-walk bootstrapping approach can be applied without human assistance directly to large planning domains. However, as already pointed out, our goal of performing well on the LRW distribution may not always correspond well with a particular target problem distribution.

---

16. The most complex blocks-world goal for RRL was to achieve $\mathbf{on}(A, B)$ in an $n$ block environment. We consider blocks-world goals that involve all $n$ blocks.

Our bootstrapping approach is similar in spirit to the bootstrapping framework of "learning from exercises" (Natarajan, 1989; Reddy & Tadepalli, 1997). Here, the learner is provided with planning problems, or exercises, in order of increasing difficulty. After learning on easier problems, the learner is able to use its new knowledge, or skills, in order to bootstrap learning on the harder problems. This work, however, has previously relied on a human to provide the exercises, which typically requires insight into the planning domain and the underlying form of control knowledge and planner. Our work can be viewed as an automatic instantiation of "learning from exercises", specifically designed for learning LRW policies.

Our random-walk bootstrapping is most similar to the approach used in Micro-Hillary (Finkelstein & Markovitch, 1998), a macro-learning system for problem solving. In that work, instead of generating problems via random walks starting at an initial state, random walks were generated backward from goal states. This approach assumes that actions are invertible or that we are given a set of "backward actions". When such assumptions hold, the backward random-walk approach may be preferable when we are provided with a goal distribution that does not match well with the goals generated by forward random walks. Of course, in other cases forward random walks may be preferable. Micro-Hillary was empirically tested in the $N \times N$ sliding-puzzle domain; however, as discussed in that work, there remain challenges for applying the system to more complex domains with parameterized actions and recursive structure, such as familiar STRIPS domains. To the best of our knowledge, the idea of learning from random walks has not been previously explored in the context of STRIPS planning domains.

The idea of searching for a good policy directly in policy space rather than value-function space is a primary motivation for policy-gradient RL algorithms. However, these algorithms have been largely explored in the context of parametric policy spaces. While this approach has demonstrated impressive success in a number of domains, it appears difficult to define such policy spaces for the types of planning problem considered here.

Our API approach can be viewed as a type of reduction from planning or reinforcement learning to classification learning. That is, we solve an MDP by generating and solving a series of cost-sensitive classification problems. Recently, there have been several other proposals for reducing reinforcement learning to classification. Dietterich and Wang (2001) proposed a reinforcement learning approach based on batch value function approximation. One of the proposed approximations enforced only that the learned approximation assign the best action the highest value, which is a type of classifier learning. Lagoudakis and Parr (2003) proposed a classification-based API approach that is closely related to ours. The primary difference is the form of the classification problem produced on each iteration. They generate standard multi-class classification problems, whereas we generate cost-sensitive problems. Bagnell, Kakade, Ng, and Schneider (2003) introduced a closely related algorithm for learning non-stationary policies in reinforcement learning. For a specified horizon time $h$, their approach learns a sequence of $h$ policies. At each iteration, all policies are held fixed except for one, which is optimized by forming a classification problem via policy rollout[17]. Finally, Langford and Zadrozny (2004) provide a formal reduction from reinforcement learning to classification, showing that $\epsilon$-accurate classification learning implies near-optimal reinforcement learning. This approach uses an optimistic variant of sparse sampling to generate $h$ classification problems, one for each horizon time step.

## 9. Summary and Future Work

We introduced a new variant of API that learns policies directly, without representing approximate value functions. This allowed us to utilize a relational policy language for learning compact policy representations. We also introduced a new API bootstrapping technique for goal-based planning domains. Our experiments show that the **LRW-API** algorithm, which combines these techniques, is able to find good policies for a variety of relational MDPs corresponding to classical planning domains and their stochastic variants. We know of no previous MDP technique that has been successfully applied to problems such as these.

Our experiments also pointed to a number of weaknesses of our current approach. First, our bootstrapping technique, based on long random walks, does not always correspond well to the problem distribution of interest. Investigating other automatic bootstrapping techniques is an interesting direction, related to the general problems of exploration and reward shaping in reinforcement learning. Second, we have seen that limitations of our current policy language and learner are partly responsible for some of the failures of our system. In such cases, we must either: 1) depend on the human to provide useful features to the system, or 2) extend the policy language and develop more advanced learning techniques. Policy-language extensions that we are considering include various extensions to the knowledge representation used to represent sets of objects in the domain (in particular, for route-finding in maps/grids), as well as non-reactive policies that incorporate search into decision-making.

As we consider ever more complex planning domains, it is inevitable that our brute-force enumeration approach to learning policies from trajectories will not scale. Presently our policy learner, as well as the entire API technique, makes no attempt to use the definition of a domain when one is available. We believe that developing a learner that can exploit this information to bias its search for good policies is an important direction of future work. Recently, Gretton and Thiebaux (2004) have taken a step in this direction by using logical regression (based on a domain model) to generate candidate rules for the learner. Developing tractable variations of this approach is a promising research direction. In addition, exploring other ways of incorporating a domain model into our approach and other model-blind approaches is critical. Ultimately, scalable AI planning systems will need to combine experience with stronger forms of explicit reasoning.

---

17. Here the initial state distribution is dictated by the policies at previous time steps, which are held fixed. Likewise the actions selected along the rollout trajectories are dictated by policies at future time steps, which are also held fixed.

## Acknowledgments

We would like to thank Lin Zhu for originally suggesting the idea of using random walks for bootstrapping. We would also like to thank the reviewers and editors for helping to vastly improve this paper. This work was supported in part by NSF grants 9977981-IIS and 0093100-IIS.

## Appendix A. Omitted Proofs

**Proposition 1.** *Let $H$ be a finite class of deterministic policies. For any $\pi \in H$, and any set of $n = \epsilon^{-1} \ln \frac{|H|}{\delta}$ trajectories drawn independently from $D_h^\pi$, there is a $1 - \delta$ probability that every $\hat{\pi} \in H$ consistent with the trajectories satisfies $\overline{V}(\hat{\pi}) \geq \overline{V}(\pi) - 2V_{max}(\epsilon + \gamma^h)$.*

**Proof**: We first introduce some basic properties and notation that will be used below. For any deterministic policy $\pi$, if $\pi$ is consistent with a trajectory $t$, then $D_h^\pi(t)$ is entirely determined by the underlying MDP transition dynamics. This implies that if two deterministic policies $\pi$ and $\pi'$ are both consistent with a trajectory $t$ then $D_h^\pi(t) = D_h^{\pi'}(t)$. We will denote by $v(t)$ the cumulative discounted reward accumulated by executing trajectory $t$. For any policy $\pi$, we have that $\overline{V}_h(\pi) = \sum_t D_h^\pi(t) \cdot v(t)$ where the summation is taken over all length $h$ trajectories (or simply those that are consistent with $\pi$). Finally for a set of trajectories $\Gamma$ we will let $D_h^\pi(\Gamma) = \sum_{t \in \Gamma'} D_h^\pi(t)$ giving the cumulative probability of $\pi$ generating the trajectories in $\Gamma$.

Consider a particular $\pi \in H$ and any $\hat{\pi} \in H$ that is consistent with the $n$ trajectories of $\pi$. We will let $\Gamma$ denote the set of all length $h$ trajectories that are consistent with $\pi$ and $\hat{\Gamma}$ denote the set of trajectories that are consistent with $\hat{\pi}$. Following Khardon (1999b) we first give a standard argument showing that with high probability $D_h^\pi(\hat{\Gamma}) > 1 - \epsilon$. To see this consider the probability that $\hat{\pi}$ is consistent with all $n = \epsilon^{-1} \ln \frac{|H|}{\delta}$ trajectories of $\pi$ given that $D_h^\pi(\hat{\Gamma}) \leq 1 - \epsilon$. The probability that this occurs is at most $(1 - \epsilon)^n < e^{-\epsilon n} = \frac{\delta}{|H|}$. Thus the probability of choosing such a $\hat{\pi}$ is at most $|H| \frac{\delta}{|H|} = \delta$. Thus, with probability at least $1 - \delta$ we know that $D_h^\pi(\hat{\Gamma}) > 1 - \epsilon$. Note that $D_h^\pi(\hat{\Gamma}) = D_h^{\hat{\pi}}(\Gamma)$.

Now given the condition that $D_h^\pi(\hat{\Gamma}) > 1 - \epsilon$ we show that $\overline{V}_h(\hat{\pi}) \geq \overline{V}_h(\pi) - 2\epsilon V_{\max}$ by considering the difference of the two value functions.

$$
\begin{aligned}
\overline{V}_h(\pi) - \overline{V}_h(\hat{\pi}) &= \sum_{t \in \Gamma} D_h^\pi(t) \cdot v(t) - \sum_{t \in \hat{\Gamma}} D_h^{\hat{\pi}}(t) \cdot v(t) \\
&= \sum_{t \in \Gamma - \hat{\Gamma}} D_h^\pi(t) \cdot v(t) + \sum_{t \in \Gamma \cap \hat{\Gamma}} (D_h^\pi(t) - D_h^{\hat{\pi}}(t)) \cdot v(t) - \sum_{t \in \hat{\Gamma} - \Gamma} D_h^{\hat{\pi}}(t) \cdot v(t) \\
&= \sum_{t \in \Gamma - \hat{\Gamma}} D_h^\pi(t) \cdot v(t) + 0 - \sum_{t \in \hat{\Gamma} - \Gamma} D_h^{\hat{\pi}}(t) \cdot v(t) \\
&\leq V_{\max}[D_h^\pi(\Gamma - \hat{\Gamma}) + D_h^{\hat{\pi}}(\hat{\Gamma} - \Gamma)] \\
&= V_{\max}[1 - D_h^\pi(\hat{\Gamma}) + 1 - D_h^{\hat{\pi}}(\Gamma)] \\
&\leq 2\epsilon V_{\max}
\end{aligned}
$$

The third lines follows since $D_h^\pi(t) = D_h^{\hat\pi}(t)$ when $\pi$ and $\hat\pi$ are both consistent with $t$. The last line follows by substituting our assumption of $D_h^\pi(\hat\Gamma) = D_h^{\hat\pi}(\Gamma) > 1-\epsilon$ into the previous line. Combining this result with the approximation due to using a finite horizon,

$$\overline{V}(\pi) - \overline{V}(\hat\pi) \le \overline{V}_h(\pi) - \overline{V}_h(\hat\pi) + 2\gamma^h V_{\max}$$

we get that with probability at least $1-\delta$, $\overline{V}(\pi) - \overline{V}(\hat\pi) \le 2V_{\max}(\epsilon + \gamma^h)$, which completes the proof. $\quad\square$

**Proposition 2.** *For any MDP with Q-advantage at least $\Delta^*$, and any $0 < \delta' < 1$, if we have*

$$
\begin{aligned}
h &> \log_\gamma \frac{\Delta^*}{8V_{max}} \\
w &> \left(\frac{8V_{max}}{\Delta^*}\right)^2 \ln\frac{|A|}{\delta'} \\
\Delta &= \frac{\Delta^*}{2}
\end{aligned}
$$

*then for any state $s$, $\hat{A}(\Delta, s) = A^\pi(s)$ with probability at least $1-\delta'$.*

**Proof:** Given a real valued random variable $X$ bounded in absolute value by $X_{\max}$ and an average $\hat{X}$ of $w$ independently drawn samples of $X$, the additive Chernoff bound states that with probability at least $1-\delta$, $|E[X] - \hat{X}| \le X_{\max}\sqrt{\frac{-\ln\delta}{w}}$.

Note that $Q_h^\pi(s, a)$ is the expectation of the random variable $X(s, a) = R(s, a) + \gamma V_{h-1}^\pi(T(s, a))$ and $\hat{Q}(s, a)$ is simply an average of $w$ independent samples of $X(s, a)$. The Chernoff bound tells us that with probability at least $1 - \frac{\delta'}{|A|}$, $|Q_h^\pi(s, a) - \hat{Q}(s, a)| \le V_{\max}\sqrt{\frac{\ln|A| - \ln\delta'}{w}}$, where $|A|$ is the number of actions. Substituting in our choice of $w$ we get that with probability at least $1 - \delta'$, $|Q_h^\pi(s, a) - \hat{Q}(s, a)| < \frac{\Delta^*}{8}$ is satisfied by all actions simultaneously. We also know that $|Q^\pi(s, a) - Q_h^\pi(s, a)| \le \gamma^h V_{\max}$, which by our choice of $h$ gives, $|Q^\pi(s, a) - Q_h^\pi(s, a)| < \frac{\Delta^*}{8}$. Combining these relationships we get that with probability at least $1 - \delta'$, $|Q^\pi(s, a) - \hat{Q}(s, a)| < \frac{\Delta^*}{4}$ holds for all actions simultaneously.

We can use this bound to show that with high probability the Q-value estimates for actions in $A^\pi(s)$ will be within a $\frac{\Delta^*}{2}$ range of each other, and other actions will be outside of that range. In particular, consider any action $a \in A^\pi(s)$ and some other action $a'$. If $a' \in A^\pi(s)$ then we have that $Q^\pi(s, a) = Q^\pi(s, a')$. From the above bound we get that $|\hat{Q}(s, a) - \hat{Q}(s, a')| < \frac{\Delta^*}{2}$. Otherwise $a' \notin A^\pi(s)$ and by our assumption about the MDP Q-advantage we get that $Q^\pi(s, a) - Q^\pi(s, a') \ge \Delta^*$. Using the above bound this implies that $\hat{Q}(s, a) - \hat{Q}(s, a') > \frac{\Delta^*}{2}$. These relationships and the definition of $\hat{A}(\Delta, s)$ imply that with probability at least $1 - \delta'$ we have that $\hat{A}(\Delta, s) = A^\pi(s)$. $\quad\square$

## Appendix B. Learned Policies

Below we give the final taxonomic-decision-list policies that were learned for each domain in our experiments. Rather than write rules in the form $a(x_1, \ldots, x_k) : L_1 \wedge L_2 \wedge \cdots \wedge L_m$ we drop the variables from the head and simply write, $a : L_1 \wedge L_2 \wedge \cdots \wedge L_m$. In addition below we use the notation $R^{-*}$ as short-hand for $(R^{-1})^*$ where $R$ is a relation. When interpreting the policies, it is important to remember that for each rule of action type $a$, the preconditions for action type $a$ are implicitly included in the constraints. Thus, the rules will often allow actions that are not legal, but those actions will never be considered by the system.

**Gripper**

1. MOVE: $(X_1 \in (\text{NOT } (\text{GAT } (\text{CARRY}^{-1} \text{ GRIPPER})))) \wedge (X_2 \in (\text{NOT } (\text{GAT } (\text{AT}^{-1} \text{ AT-ROBBY})))) \wedge (X_2 \in (\text{GAT } (\text{NOT } (\text{CAT}^{-1} \text{ ROOM})))) \wedge (X_1 \in (\text{CAT BALL}))$

2. DROP: $(X_1 \in (\text{GAT}^{-1} \text{ AT-ROBBY}))$

3. PICK: $(X_1 \in (\text{GAT}^{-1} (\text{GAT } (\text{CARRY}^{-1} \text{ GRIPPER})))) \wedge (X_1 \in (\text{GAT}^{-1} (\text{NOT AT-ROBBY})))$

4. PICK: $(X_2 \in (\text{AT } (\text{NOT } (\text{GAT}^{-1} \text{ ROOM})))) \wedge (X_1 \in (\text{GAT}^{-1} (\text{NOT AT-ROBBY})))$

5. PICK: $(X_1 \in (\text{GAT}^{-1} (\text{NOT AT-ROBBY})))$

**Briefcase**

1. PUT-IN: $(X_1 \in (\text{GAT}^{-1} (\text{NOT IS-AT})))$

2. MOVE: $(X_2 \in (\text{AT } (\text{NOT } (\text{CAT}^{-1} \text{ LOCATION})))) \wedge (X_2 \in (\text{NOT } (\text{AT } (\text{GAT}^{-1} \text{ CIS-AT}))))$

3. MOVE: $(X_2 \in (\text{GAT IN})) \wedge (X_1 \in (\text{NOT } (\text{CAT IN})))$

4. TAKE-OUT: $(X_1 \in (\text{CAT}^{-1} \text{ IS-AT}))$

5. MOVE: $(X_2 \in \text{GIS-AT})$

6. MOVE: $(X_2 \in (\text{AT } (\text{GAT}^{-1} \text{ CIS-AT})))$

7. PUT-IN: $(X_1 \in \text{UNIVERSAL})$

**Schedule**

1. DO-IMMERSION-PAINT: $(X_1 \in (\text{NOT } (\text{PAINTED}^{-1} X_2))) \wedge (X_1 \in (\text{GPAINTED}^{-1} X_2))$

2. DO-DRILL-PRESS: $(X_1 \in (\text{GHAS-HOLEO}^{-1} X_3)) \wedge (X_1 \in (\text{GHAS-HOLEW}^{-1} X_2))$

3. DO-LATHE: $(X_1 \in (\text{NOT } (\text{SHAPE}^{-1} \text{ CYLINDRICAL}))) \wedge (X_1 \in (\text{GSHAPE}^{-1} \text{ CYLINDRICAL}))$

4. DO-DRILL-PRESS: $(X_1 \in (\text{GHAS-HOLEW}^{-1} X_2))$

5. DO-DRILL-PRESS: $(X_1 \in (\text{GHAS-HOLEO}^{-1} X_3))$

6. DO-GRIND: $(X_1 \in (\text{NOT } (\text{SURFACE-CONDITION}^{-1} \text{ SMOOTH}))) \wedge (X_1 \in (\text{GSURFACE-CONDITION}^{-1} \text{ SMOOTH}))$

7. DO-POLISH: $(X_1 \in (\text{NOT } (\text{SURFACE-CONDITION}^{-1} \text{ POLISHED}))) \wedge (X_1 \in (\text{GSURFACE-CONDITION}^{-1} \text{ POLISHED}))$

8. DO-TIME-STEP:

**Elevator**

1. DEPART: $(X_2 \in \text{GSERVED})$

2. DOWN: $(X_2 \in (\text{DESTIN BOARDED})) \wedge (X_2 \in (\text{DESTIN GSERVED}))$

3. UP: $(X_2 \in (\text{DESTIN BOARDED})) \wedge (X_2 \in (\text{DESTIN GSERVED})) \wedge (X_2 \in (\text{ABOVE } (\text{ORIGIN BOARDED}))) \wedge (X_1 \text{ (NOT } (\text{DESTIN BOARDED})))$

4. BOARD: $(X_2 \in (\text{NOT CSERVED})) \wedge (X_2 \in \text{GSERVED})$

5. UP: $(X_2 \in (\text{ORIGIN GSERVED})) \wedge (X_2 \in (\text{NOT } (\text{DESTIN BOARDED}))) \wedge (X_2 \in (\text{NOT } (\text{DESTIN GSERVED}))) \wedge (X_2 \in (\text{ORIGIN } (\text{NOT CSERVED}))) \wedge (X_2 \in (\text{ABOVE } (\text{DESTIN PASSENGER}))) \wedge (X_1 \in (\text{NOT } (\text{DESTIN BOARDED})))$

6. DOWN: $(X_2 \in (\text{ORIGIN GSERVED})) \wedge (X_2 \in (\text{ORIGIN } (\text{NOT CSERVED}))) \wedge (X_1 \in (\text{NOT } (\text{DESTIN BOARDED})))$

7. UP: $(X_2 \in (\text{NOT (ORIGIN BOARDED)})) \wedge (X_2 \in (\text{NOT (DESTIN BOARDED)}))$

**FreeCell**

1. SENDTOHOME: $(X_1 (\text{CANSTACK}^{-1} (\text{CANSTACK (SUIT}^{-1} (\text{SUIT INCELL}))))) \wedge (X_5 \in (\text{NOT GHOME}))$

2. MOVE-B: $(X_2 \in (\text{NOT (CANSTACK (ON GHOME)}))) \wedge (X_2 \in (\text{CANSTACK GHOME})) \wedge (X_2 \in (\text{VALUE}^{-1} (\text{NOT COLSPACE}))) \wedge (X_1 \in (\text{CANSTACK}^{-1} (\text{SUIT}^{-1} (\text{SUIT BOTTOMCOL}))))$

3. MOVE: $(X_1 \in (\text{CANSTACK}^{-1} (\text{ON (CANSTACK}^{-1} (\text{ON}^{-1} \text{ GHOME}))))) \wedge (X_3 \in (\text{CANSTACK (ON (SUIT}^{-1} (\text{SUIT BOTTOMCOL})))) \wedge (X_1 \in (\text{ON}^{-1} \text{ BOTTOMCOL})) \wedge (X_1 \in (\text{CANSTACK}^{-1} (\text{ON}^{-1} (\text{NOT (CANSTACK (VALUE}^{-1} \text{ CELLSPACE}))))))) \wedge (X_1 \in (\text{NOT (CANSTACK}^{-1} (\text{SUIT}^{-1} (\text{SUIT INCELL})))) \wedge (X_3 \in (\text{CANSTACK BOTTOMCOL})) \wedge (X_1 \in (\text{SUIT}^{-1} (\text{SUIT (ON}^{-1} (\text{NOT (CANSTACK (VALUE}^{-1} \text{ CELLSPACE}))))))) \wedge (X_1 \in (\text{VALUE}^{-1} (\text{NOT COLSPACE})))$ $(0 \ (\text{ON}^{-1} (\text{NOT (CANSTACK}^{-1} (\text{SUIT}^{-1} (\text{SUIT INCELL}))))) \wedge (X_1 \in (\text{NOT (CANSTACK}^{-1} \text{ CHOME})))$

4. SENDTOHOME-B: $(X_4 \in (\text{NOT GHOME}))$

5. SENDTOHOME: $(X_1 \in (\text{ON}^{-1} (\text{CANSTACK (CANSTACK}^{-1} (\text{SUIT}^{-1} (\text{SUIT INCELL})))))) \wedge (X_5 \in (\text{NOT GHOME}))$

6. SENDTOHOME: $(X_1 (\text{ON}^{-1} (\text{ON}^{-1} \text{ GHOME}))) \wedge (X_1 \in (\text{CANSTACK}^{-1} (\text{NOT GHOME}))) \wedge (X_1 \in (\text{CANSTACK}^{-1} (\text{ON}^{-1} \text{ GHOME}))))$ $(X_5 \in (\text{NOT GHOME}))$

7. MOVE-B: $(X_1 \in (\text{NOT (CANSTACK}^{-1} \text{ GHOME}))) \wedge (X_2 \in (\text{VALUE}^{-1} (\text{NOT COLSPACE}))) \wedge (X_1 \in (\text{CANSTACK}^{-1} (\text{SUIT}^{-1} (\text{SUIT BOTTOMCOL}))))$

8. SENDTOFREE: $(X_1 \in (\text{ON}^{-1} (\text{ON}^{-1} \text{ GHOME}))) \wedge (X_1 \in (\text{NOT GHOME}))$

9. SENDTOHOME: $(X_5 \in (\text{CANSTACK}^{-1} (\text{CANSTACK (ON GHOME}))) \wedge (X_5 \in (\text{NOT GHOME}))$

10. SENDTOHOME: $(0 \text{ GHOME}) \ (X_5 \in (\text{VALUE}^{-1} (\text{NOT COLSPACE}))) \wedge (X_5 \in (\text{NOT (CANSTACK}^{-1} (\text{ON}^{-1} (\text{NOT GHOME}))))) \wedge (X_1 \in (\text{ON}^{-1} (\text{NOT (ON}^{-1} \text{ GHOME}))) \wedge (X_5 \in (\text{NOT GHOME}))$

11. NEWCOLFROMFREECELL: $(X_1 \in \text{GHOME})$

12. SENDTOHOME: $(X_5 \in (\text{CANSTACK}^{-1} (\text{ON GHOME}))) \wedge (X_1 \in \text{GHOME}) \wedge (X_5 \in (\text{NOT GHOME}))$

13. MOVE-B: $(X_1 \in (\text{VALUE}^{-1} (\text{VALUE HOME}))) \wedge (X_2 \in (\text{VALUE}^{-1} (\text{NOT COLSPACE}))) \wedge (X_1 \in (\text{CANSTACK}^{-1} (\text{SUIT}^{-1} (\text{SUIT BOTTOMCOL}))))$

14. SENDTOHOME: $(X_1 \in (\text{CANSTACK}^{-1} (\text{ON}^{-1} (\text{CANSTACK}^{-1} (\text{SUIT}^{-1} (\text{SUIT INCELL})))))) \wedge (X_5 \in (\text{NOT GHOME}))$

15. SENDTOHOME: $(X_1 \in (\text{ON}^{-1} (\text{ON}^{-1} (\text{CANSTACK}^{-1} (\text{ON}^{-1} (\text{NOT GHOME})))))) \wedge (X_5 \in (\text{NOT GHOME}))$

16. SENDTOFREE: $(X_1 \in (\text{CANSTACK}^{-1} (\text{ON (ON}^{-1} \text{ GHOME})))) \wedge (X_1 \in (\text{SUIT}^{-1} (\text{SUIT BOTTOMCOL}))) \wedge (X_1 \in (\text{ON}^{-1} \text{ BOTTOMCOL}))$

17. MOVE: $(X_3 \in (\text{ON}^{-1} (\text{CANSTACK}^{-1} \text{ CLEAR}))) \wedge (X_1 \in (\text{ON}^{-1} (\text{CANSTACK (ON}^{-1} (\text{NOT (CANSTACK (VALUE}^{-1} \text{ CELLSPACE}))))))) \wedge (X_3 \in (\text{NOT GHOME})) \wedge (X_1 \in (\text{CANSTACK (ON}^{-1} (\text{NOT (CANSTACK (VALUE}^{-1} \text{ CELLSPACE})))))) \wedge (X_1 \in (\text{NOT (CANSTACK}^{-1} (\text{SUIT}^{-1} (\text{SUIT INCELL}))))) \wedge (X_1 \in (\text{ON}^{-1} \text{ BOTTOMCOL})) \wedge (X_1 \in (\text{SUIT}^{-1} (\text{SUIT (ON}^{-1} (\text{NOT (CANSTACK (VALUE}^{-1} \text{ CELLSPACE})))))))) \wedge (X_1 \in (\text{VALUE}^{-1} (\text{NOT COLSPACE}))) \wedge (X_1 \in (\text{ON}^{-1} (\text{NOT (CANSTACK}^{-1} (\text{SUIT}^{-1} (\text{SUIT INCELL})))))) \wedge (X_1 \in (\text{NOT (CANSTACK}^{-1} \text{ CHOME})))$

18. MOVE: $(X_1 \in (\text{SUIT}^{-1} (\text{SUIT CHOME}))) \wedge (X_3 \in (\text{NOT GHOME})) \wedge (X_3 \in (\text{NOT (ON}^{-1} \text{ GHOME}))) \wedge (X_1 \in (\text{ON}^{-1} (\text{CANSTACK}^{-1} \text{ BOTTOMCOL})))$

19. SENDTOHOME: $(X_1 \in (\text{CANSTACK (ON (CANSTACK (ON GHOME}))))) \wedge (X_1 \in \text{GHOME}) \ (X_5 \in (\text{NOT GHOME}))$

20. SENDTOHOME: $(X_1 \in (\text{CANSTACK}^{-1} (\text{ON (CANSTACK}^{-1} (\text{ON}^{-1} \text{ GHOME}))))) \wedge (X_1 \in (\text{NOT (SUIT}^{-1} (\text{SUIT BOTTOMCOL})))) \wedge (X_5 \in (\text{NOT GHOME}))$

21. SENDTOFREE: $(X_1 \in (\text{CANSTACK (ON (CANSTACK (VALUE}^{-1} \text{ CELLSPACE}))))) \wedge (X_1 \in (\text{CANSTACK CHOME}))$

22. SENDTOHOME: $(X_1 \in (\text{CANSTACK}^{-1} (\text{SUIT}^{-1} (\text{SUIT INCELL})))) \wedge (X_1 \in (\text{ON}^{-1} (\text{NOT (CANSTACK (VALUE}^{-1} \text{ CELLSPACE}))))) \wedge (X_5 \in (\text{NOT GHOME}))$

23. SENDTONEWCOL: $(X_1 \in (\text{CANSTACK (CANSTACK}^{-1} (\text{ON}^{-1} \text{ GHOME}))))$

24. SENDTOFREE: $(X_1 \in (\text{CANSTACK (ON}^{-1} (\text{CANSTACK}^{-1} (\text{ON}^{-1} \text{ GHOME}))))) \wedge (X_1 \in (\text{NOT (CANSTACK GHOME}))) \wedge (X_1 \in (\text{NOT (ON}^{-1} \text{ GHOME}))) \wedge (X_1 \in (\text{ON}^{-1} (\text{NOT (CANSTACK}^{-1} (\text{SUIT}^{-1} (\text{SUIT INCELL}))))))$

25. SENDTOFREE: $(X_1 \in (\text{ON}^{-1} (\text{CANSTACK} (\text{CANSTACK}^{-1} (\text{ON}^{-1} \text{ GHOME}))))) \wedge (X_1 \in (\text{NOT} (\text{CANSTACK BOTTOM-COL}))) \wedge (X_1 \in (\text{NOT} (\text{CANSTACK}^{-1} (\text{CANSTACK} (\text{ON GHOME})))))$

26. SENDTOFREE: $(X_1 \in (\text{CANSTACK} (\text{ON}^{-1} (\text{CANSTACK}^{-1} (\text{ON}^{-1} (\text{NOT GHOME})))))) \wedge (X_1 \in (\text{NOT} (\text{CANSTACK GHOME}))) \wedge (X_1 \in (\text{CANSTACK} (\text{NOT} (\text{SUIT}^{-1} (\text{SUIT BOTTOMCOL})))))$

27. SENDTOHOME: $(X_1 \in (\text{CANSTACK}^{-1} (\text{CANSTACK} (\text{ON}^{-1} \text{ GHOME})))) \wedge (X_1 \in (\text{ON}^{-1} (\text{CANSTACK}^{-1} (\text{ON}^{-1} (\text{NOT GHOME})))) \wedge (X_1 \in (\text{NOT GHOME})) \wedge (X_5 \in (\text{NOT GHOME}))$

28. SENDTOFREE: $(X_1 \in (\text{CANSTACK} (\text{ON}^{-1} (\text{CANSTACK}^{-1} (\text{ON}^{-1} (\text{NOT GHOME})))))) \wedge (X_1 \in (\text{CANSTACK} (\text{CANSTACK}^{-1} (\text{ON}^{-1} \text{ GHOME})))) \wedge (X_1 \in (\text{NOT GHOME})) \wedge (X_1 \in (\text{ON}^{-1} (\text{CANSTACK}^{-1} (\text{ON}^{-1} (\text{NOT} (\text{CANSTACK} (\text{VALUE}^{-1} \text{ CELLSPACE}))))))))$

29. SENDTOFREE: $(X_1 \in (\text{CANSTACK CHOME})) \wedge (X_1 \in (\text{SUIT}^{-1} (\text{SUIT} (\text{CANSTACK}^{-1} (\text{ON}^{-1} \text{ GHOME})))))$

30. SENDTOHOME: $(X_1 \in \text{GHOME}) \wedge (X_1 \in (\text{SUIT}^{-1} (\text{SUIT BOTTOMCOL}))) \wedge (X_1 \in (\text{CANSTACK}^{-1} (\text{NOT} (\text{ON}^{-1} \text{ GHOME})))) \wedge (X_5 \in (\text{NOT GHOME}))$

31. SENDTOFREE: $(X_1 \in (\text{CANSTACK}^{-1} (\text{ON}^{-1} \text{ GHOME}))) \wedge (X_1 \in (\text{CANSTACK}^{-1} (\text{ON}^{-1} (\text{NOT GHOME}))))$

32. SENDTOFREE: $(X_1 \in (\text{CANSTACK} (\text{ON}^{-1} \text{ GHOME}))) \wedge (X_1 \in (\text{NOT GHOME})) \wedge (X_1 \in (\text{ON}^{-1} (\text{CANSTACK}^{-1} (\text{ON}^{-1} (\text{NOT GHOME})))))$

33. SENDTOHOME: $(X_1 \in (\text{ON}^{-1} (\text{CANSTACK}^{-1} \text{ BOTTOMCOL}))) \wedge (X_1 \in (\text{CANSTACK}^{-1} (\text{NOT GHOME}))) \wedge (X_5 \in (\text{NOT GHOME}))$

34. SENDTOFREE: $(X_1 \in (\text{CANSTACK} (\text{ON} (\text{CANSTACK}^{-1} (\text{ON}^{-1} (\text{NOT GHOME})))))) \wedge (X_1 \in (\text{NOT} (\text{SUIT}^{-1} (\text{SUIT BOTTOMCOL})))) \wedge (X_1 \in (\text{NOT GHOME}))$

35. SENDTOHOME: $(X_1 \in (\text{NOT} (\text{CANSTACK}^{-1} \text{ GHOME}))) \wedge (X_1 \in (\text{NOT} (\text{SUIT}^{-1} (\text{SUIT BOTTOMCOL})))) \wedge (X_5 \in (\text{NOT GHOME}))$

36. SENDTOFREE: $(X_1 \in (\text{NOT} (\text{ON}^{-1} \text{ GHOME}))) \wedge (X_1 \in (\text{CANSTACK} (\text{CANSTACK}^{-1} (\text{ON}^{-1} (\text{NOT GHOME})))))$

37. SENDTOFREE-B: $(X_1 \in (\text{NOT GHOME}))$

38. SENDTOFREE: $(X_1 \in \text{UNIVERSAL})$

### Logistics

1. FLY-AIRPLANE: $(X_1 \in (\text{IN} (\text{GAT}^{-1} \text{ AIRPORT}))) \wedge (X_1 \in (\text{NOT} (\text{IN} (\text{GAT}^{-1} (\text{AT AIRPLANE}))))) \wedge (X_3 \in (\text{NOT} (\text{GAT} (\text{IN}^{-1} \text{ TRUCK}))))) \wedge (X_1 \in (\text{NOT} (\text{IN} (\text{GAT}^{-1} (\text{NOT AIRPORT}))))$

2. LOAD-TRUCK: $(X_2 \in (\text{IN} (\text{NOT} (\text{GAT}^{-1} (\text{NOT AIRPORT}))))) \wedge (X_1 \in (\text{GAT}^{-1} (\text{GAT} (\text{IN}^{-1} \text{ TRUCK})))) \wedge (X_1 \in (\text{NOT} (\text{CAT}^{-1} \text{ LOCATION})))$

3. DRIVE-TRUCK: $(X_3 \in (\text{AT} (\text{AT}^{-1} (\text{GAT} (\text{IN}^{-1} \text{ TRUCK}))))) \wedge (X_3 \in (\text{IN-CITY}^{-1} (\text{IN-CITY} (\text{AT AIRPLANE})))) \wedge (X_1 \in (\text{AT}^{-1} (\text{NOT} (\text{GAT} (\text{IN}^{-1} \text{ TRUCK})))))$

4. UNLOAD-TRUCK: $(X_1 \in (\text{GAT}^{-1} (\text{AT} (\text{IN OBJ})))) \wedge (X_1 \in (\text{GAT}^{-1} (\text{AT OBJ}))) \wedge (X_1 \in (\text{NOT} (\text{GAT}^{-1} (\text{AT AIRPLANE})))) \wedge (X_2 \in (\text{AT}^{-1} (\text{GAT} (\text{IN}^{-1} \text{ TRUCK})))) \wedge (X_1 \in (\text{GAT}^{-1} (\text{AT TRUCK})))$

5. FLY-AIRPLANE: $(X_3 \in (\text{GAT} (\text{IN}^{-1} \text{ AIRPLANE}))) \wedge (X_1 \in (\text{IN} (\text{NOT} (\text{GAT}^{-1} (\text{AT TRUCK}))))) \wedge (X_1 \in (\text{AT}^{-1} (\text{NOT} (\text{GAT} (\text{IN}^{-1} \text{ TRUCK})))))$

6. UNLOAD-AIRPLANE: $(X_2 \in (\text{NOT} (\text{IN} (\text{GAT}^{-1} (\text{NOT AIRPORT}))))) \wedge (X_1 \in (\text{GAT}^{-1} (\text{AT AIRPLANE})))$

7. LOAD-TRUCK: $(X_2 \in (\text{IN} (\text{NOT} (\text{GAT}^{-1} \text{ LOCATION})))) \wedge (X_1 \in (\text{NOT} (\text{GAT}^{-1} (\text{AT TRUCK})))) \wedge (X_1 \in (\text{GAT}^{-1} \text{ LOCATION}))$

8. UNLOAD-TRUCK: $(X_1 \in (\text{GAT}^{-1} (\text{AT TRUCK}))) \wedge (X_2 \in (\text{AT}^{-1} \text{ AIRPORT})) \wedge (X_2 \in (\text{NOT} (\text{IN} (\text{GAT}^{-1} (\text{NOT AIRPORT}))))) \wedge (X_1 \in (\text{GAT}^{-1} (\text{AT AIRPLANE})))$

9. FLY-AIRPLANE: $(X_3 \in (\text{AT} (\text{AT}^{-1} (\text{GAT} (\text{IN}^{-1} \text{ TRUCK}))))) \wedge (X_1 \in (\text{AT}^{-1} (\text{GAT} (\text{GAT}^{-1} \text{ LOCATION})))) \wedge (X_1 \in (\text{NOT} (\text{AT}^{-1} (\text{CAT OBJ}))))$

10. DRIVE-TRUCK: $(X_1 \in (\text{IN} (\text{GAT}^{-1} \text{ LOCATION}))) \wedge (X_1 \in (\text{AT}^{-1} (\text{NOT} (\text{GAT} (\text{IN}^{-1} \text{ TRUCK}))))) \wedge (X_1 \in (\text{AT}^{-1} (\text{NOT} (\text{AT AIRPLANE}))))$

11. UNLOAD-TRUCK: $(X_2 \in (\text{AT}^{-1} (\text{GAT} (\text{GAT}^{-1} (\text{NOT AIRPORT}))))) \wedge (X_1 \in (\text{NOT} (\text{GAT}^{-1} \text{ AIRPORT})))$

12. FLY-AIRPLANE: $(X_3 \in (\text{NOT} (\text{GAT} (\text{GAT}^{-1} \text{ LOCATION})))) \wedge (X_1 \in (\text{AT}^{-1} (\text{GAT} (\text{AT}^{-1} (\text{CAT OBJ}))))) \wedge (X_3 \in (\text{AT} (\text{NOT} (\text{GAT}^{-1} (\text{AT AIRPLANE}))))) \wedge (X_3 \in (\text{AT OBJ})) \wedge (X_1 \in (\text{NOT} (\text{IN} (\text{GAT}^{-1} \text{ AIRPORT})))) \wedge (X_3 \in (\text{NOT} (\text{AT} (\text{IN OBJ}))))$

13. UNLOAD-TRUCK: $(X_1 \in (GAT^{-1} \; AIRPORT))$

14. LOAD-TRUCK: $(X_1 \in (AT^{-1} \; (CAT \; (GAT^{-1} \; (AT \; AIRPLANE))))) \wedge (X_1 \in (NOT \; (GAT^{-1} \; LOCATION)))$

15. LOAD-TRUCK: $(X_1 \in (GAT^{-1} \; (CAT \; (GAT^{-1} \; (AT \; AIRPLANE))))) \wedge (X_1 \in (NOT \; (GAT^{-1} \; (AT \; TRUCK)))) \wedge (X_1 \in (GAT^{-1} \; (AT \; (GAT^{-1} \; (AT \; AIRPLANE)))))$

16. LOAD-TRUCK: $(X_1 \in (GAT^{-1} \; (NOT \; AIRPORT))) \wedge (X_1 \in (NOT \; (GAT^{-1} \; (AT \; TRUCK))))$

17. FLY-AIRPLANE: $(X_3 \in (AT \; (GAT^{-1} \; (AT \; AIRPLANE)))) \wedge (X_1 \in (AT^{-1} \; (CAT \; OBJ)))$

18. FLY-AIRPLANE: $(X_3 \in (NOT \; (GAT \; (AT^{-1} \; (CAT \; OBJ))))) \wedge (X_1 \in (AT^{-1} \; (GAT \; (AT^{-1} \; (CAT \; OBJ))))) \wedge (X_1 \in (AT^{-1} \; (GAT \; (GAT^{-1} \; (AT \; TRUCK)))))$

19. LOAD-TRUCK: $(X_1 \in (GAT^{-1} \; (AT \; AIRPLANE))) \wedge (X_1 \in (NOT \; (GAT^{-1} \; (AT \; TRUCK)))) \wedge (X_1 \in (AT^{-1} \; (CAT \; OBJ)))$

20. LOAD-AIRPLANE: $(X_1 \in (GAT^{-1} \; AIRPORT)) \wedge (X_1 \in (NOT \; (CAT^{-1} \; LOCATION))) \wedge (X_1 \in (GAT^{-1} \; (NOT \; (AT \; AIRPLANE)))) \wedge (X_2 \in (NOT \; (IN \; (GAT^{-1} \; (NOT \; AIRPORT)))))$

21. FLY-AIRPLANE: $(X_3 \in (AT \; (GAT^{-1} \; (AT \; AIRPLANE)))) \wedge (X_3 \in (NOT \; (AT \; TRUCK)))$

22. LOAD-TRUCK: $(X_1 \in (AT^{-1} \; (CAT \; (GAT^{-1} \; (NOT \; AIRPORT))))) \wedge (X_1 \in (GAT^{-1} \; AIRPORT))$

23. DRIVE-TRUCK: $(X_3 \in (NOT \; (AT \; OBJ))) \wedge (X_1 \in (NOT \; (AT^{-1} \; (CAT \; OBJ)))) \wedge (X_1 \in (AT^{-1} \; (GAT \; (GAT^{-1} \; LOCATION))))$

24. LOAD-TRUCK: $(X_1 \in (GAT^{-1} \; (CAT \; (CAT^{-1} \; AIRPORT)))) \wedge (X_1 \in (NOT \; (CAT^{-1} \; LOCATION)))$

25. FLY-AIRPLANE: $(X_3 \in (AT \; (GAT^{-1} \; (AT \; AIRPLANE)))) \wedge (X_1 \in (AT^{-1} \; (AT \; OBJ)))$

26. DRIVE-TRUCK: $(X_1 \in (IN \; OBJ))$

27. DRIVE-TRUCK: $(X_1 \in (AT^{-1} \; (GAT \; (GAT^{-1} \; AIRPORT)))) \wedge (X_3 \in (AT \; (GAT^{-1} \; AIRPORT))) \wedge (X_1 \in (AT^{-1} \; (NOT \; (AT \; AIRPLANE))))$

28. FLY-AIRPLANE: $(X_3 \in (CAT \; (GAT^{-1} \; (AT \; TRUCK)))) \wedge (X_1 \in (AT^{-1} \; (GAT \; (GAT^{-1} \; LOCATION))))$

29. LOAD-TRUCK: $(X_1 \in (GAT^{-1} \; (AT \; OBJ))) \wedge (X_1 \in (NOT \; (CAT^{-1} \; LOCATION)))$

30. DRIVE-TRUCK: $(X_3 \in (AT \; (GAT^{-1} \; (AT \; AIRPLANE)))) \wedge (X_1 \in (NOT \; (AT^{-1} \; (CAT \; OBJ))))$

31. DRIVE-TRUCK: $(X_3 \in (AT \; AIRPLANE)) \wedge (X_3 \in (AT \; (GAT^{-1} \; (AT \; TRUCK))))$

32. UNLOAD-AIRPLANE: $(X_2 \in (NOT \; (AT^{-1} \; (CAT \; OBJ)))) \wedge (X_1 \in (GAT^{-1} \; (NOT \; AIRPORT)))$

33. DRIVE-TRUCK: $(X_3 \in (AT \; (GAT^{-1} \; (AT \; TRUCK))))$

34. LOAD-TRUCK: $(X_1 \in (AT^{-1} \; (NOT \; AIRPORT))) \wedge (X_1 \in (GAT^{-1} \; AIRPORT))$

35. FLY-AIRPLANE: $(X_3 \in (AT \; (GAT^{-1} \; LOCATION)))$

36. FLY-AIRPLANE: $(X_1 \in (IN \; OBJ)) \wedge (X_3 \in (NOT \; (GAT \; (GAT^{-1} \; LOCATION)))) \wedge (X_1 \in (NOT \; (IN \; (GAT^{-1} \; AIRPORT)))) \wedge (X_3 \in (NOT \; (AT \; (IN \; OBJ)))) \wedge (X_1 \in (AT^{-1} \; (GAT \; (AT^{-1} \; (CAT \; OBJ)))))$

37. DRIVE-TRUCK: $(X_1 \in (AT^{-1} \; (AT \; AIRPLANE)))$

38. LOAD-AIRPLANE: $(X_1 \in (GAT^{-1} \; (NOT \; AIRPORT)))$

**Blocks World**

1. STACK: $(X_2 \in (GON \; HOLDING)) \wedge (X_2 \in (CON^{-*} \; (MIN \; GON))) \wedge (X_1 \in (GON^{-*} \; ON\text{-}TABLE))$

2. PUTDOWN:

3. UNSTACK: $(X_1 \in (ON^{-*} \; (ON \; (MIN \; GON)))) \wedge (X_2 \in (CON^{-*} \; (ON^* \; (MIN \; GON))))$

4. UNSTACK: $(X_2 \in (ON^{-1} \; (GON \; CLEAR))) \wedge (X_2 \in (GON^* \; (ON^{-*} \; (MIN \; GON)))) \wedge (X_1 \in (ON^{-*} \; (GON \; ON\text{-}TABLE))) (X_1 \in (GON^{-*} \; (NOT \; CLEAR)))$

5. PICKUP: $(X_1 \in (GON^{-1} \; (CON^{-*} \; (MIN \; GON)))) \wedge (X_1 \in (GON^{-1} \; CLEAR)) (X_1 \in (GON^{-1} \; (CON^{-*} \; ON\text{-}TABLE)))$

6. UNSTACK: $(X_2 \in (CON^{-*} \; (GON^{-1} \; CLEAR))) \wedge (X_1 \in (GON^{-1} \; (ON^{-*} \; (MIN \; GON)))) \wedge (X_1 \in (GON^{-1} \; (CON^* \; CLEAR)))$

7. UNSTACK: $(X_1 \in (\text{NOT } (\text{GON}^{-*} (\text{MIN GON}))))$

8. UNSTACK: $(X_2 \in (\text{GON ON-TABLE})) \wedge (X_1 \in (\text{GON}^{-1} (\text{CON}^{-*} (\text{MIN GON})))) \wedge (X_1 \in (\text{GON}^{-1} \text{ CLEAR}))$

9. UNSTACK: $(X_1 \in (\text{NOT } (\text{CON}^{-*} (\text{MIN GON})))) \wedge (X_2 \in (\text{ON}^{-*} (\text{GON}^{-1} \text{ ON-TABLE}))) \wedge (X_2 \in (\text{GON}^{-*} (\text{NOT ON-TABLE}))) \wedge (X_1 \in (\text{GON}^* (\text{GON}^{-*} \text{ ON-TABLE}))) \ (X_1 \in (\text{GON}^{-*} (\text{NOT CLEAR})))$

10. UNSTACK: $(X_2 \in (\text{NOT } (\text{CON CLEAR}))) \wedge (X_1 \in (\text{GON}^{-1} (\text{CON}^{-*} \text{ ON-TABLE})))$

11. UNSTACK: $(X_1 \in (\text{GON}^{-1} \text{ CLEAR})) \wedge (X_1 \in (\text{ON}^{-*} (\text{ON } (\text{MIN GON}))))$

**Ground Logistics**

1. LOAD: $(X_2 \in (\text{NOT } (\text{IN } (\text{GIN}^{-1} \text{ CITY})))) \wedge (X_1 \in (\text{NOT } (\text{CIN}^{-1} \text{ CITY}))) \wedge (X_1 \in (\text{GIN}^{-1} \text{ CITY}))$

2. UNLOAD: $(X_1 \in (\text{GIN}^{-1} \ X_3))$

3. DRIVE: $(X_1 \in (\text{IN } (\text{GIN}^{-1} \ X_3)))$

4. DRIVE: $(X_3 \in (\text{NOT } (\text{GIN BLOCK}))) \wedge (X_3 \in (\text{IN } (\text{GIN}^{-1} \text{ CITY}))) \wedge (X_1 \in \text{CAR}) \ (X_2 \in \text{CLEAR})$

5. DRIVE: $(X_3 \in (\text{IN } (\text{GIN}^{-1} \text{ RAIN}))) \wedge (X_1 \in \text{TRUCK})$

**Colored Blocks World**

1. PICK-UP-BLOCK-FROM: $(X_2 \in (\text{NOT } (\text{CON-TOP-OF}^{-*} \text{ TABLE}))) \wedge (X_2 \in (\text{GON-TOP-OF}^{-1} (\text{ON-TOP-OF BLOCK})))$

2. PUT-DOWN-BLOCK-ON: $(X_2 \in (\text{CON-TOP-OF}^{-1} (\text{CON-TOP-OF}^{-1} \text{ BLOCK}))) \wedge (X_2 \in (\text{GON-TOP-OF HOLDING})) \wedge (X_2 \in (\text{CON-TOP-OF}^{-*} \text{ TABLE}))$

3. PICK-UP-BLOCK-FROM: $(X_2 \in (\text{NOT } (\text{CON-TOP-OF BLOCK}))) \wedge (X_1 \in (\text{ON-TOP-OF}^{-*} (\text{CON-TOP-OF}^{-1} \text{ TABLE}))) \wedge (X_2 \in (\text{GON-TOP-OF}^* (\text{CON-TOP-OF}^{-1} \text{ BLOCK}))) \wedge (X_2 \in (\text{NOT } (\text{CON-TOP-OF}^{-1} \text{ BLOCK}))) \wedge (X_2 \in (\text{ON-TOP-OF}^{-1} (\text{GON-TOP-OF BLOCK}))) \wedge (X_1 \in (\text{GON-TOP-OF}^* (\text{GON-TOP-OF}^{-1} \text{ BLOCK})))$

4. PICK-UP-BLOCK-FROM: $(X_1 \in (\text{NOT } (\text{CON-TOP-OF}^{-*} \text{ TABLE}))) \wedge (X_1 \in (\text{GON-TOP-OF}^{-1} (\text{CON-TOP-OF}^{-*} \text{ TABLE}))) \wedge (X_1 \in (\text{GON-TOP-OF}^{-*} (\text{ON-TOP-OF}^{-1} \text{ BLOCK})))$

5. PUT-DOWN-BLOCK-ON: $(X_2 \in (\text{CON-TOP-OF}^{-1} (\text{ON-TOP-OF}^{-1} \text{ TABLE}))) \wedge (X_2 \in (\text{GON-TOP-OF HOLDING})) \wedge (X_2 \in (\text{CON-TOP-OF}^{-*} \text{ TABLE}))$

6. PUT-DOWN-BLOCK-ON: $(X_2 \in (\text{CON-TOP-OF } (\text{ON-TOP-OF BLOCK}))) \wedge (X_1 \in (\text{GON-TOP-OF}^{-1} (\text{GON-TOP-OF}^{-1} \text{ BLOCK})))$

7. PUT-DOWN-BLOCK-ON: $(X_2 \in (\text{GON-TOP-OF HOLDING})) \wedge (X_2 \in (\text{CON-TOP-OF}^{-*} \text{ TABLE}))$

8. PUT-DOWN-BLOCK-ON: $(X_2 \in \text{TABLE})$

9. PICK-UP-BLOCK-FROM: $(X_2 \in (\text{NOT } (\text{CON-TOP-OF}^{-*} \text{ TABLE}))) \wedge (X_2 \in (\text{GON-TOP-OF}^{-1} (\text{CON-TOP-OF}^{-*} \text{ TABLE})))$

10. PICK-UP-BLOCK-FROM: $(X_1 \in (\text{GON-TOP-OF}^{-1} (\text{CON-TOP-OF}^{-1} \text{ TABLE}))) \wedge (X_2 \in \text{TABLE}) \wedge (X_1 \in (\text{GON-TOP-OF } (\text{GON-TOP-OF BLOCK}))) \wedge (X_1 \in (\text{GON-TOP-OF } (\text{ON-TOP-OF}^{-1} \text{ TABLE})))$

11. PICK-UP-BLOCK-FROM: $(X_2 \in (\text{ON-TOP-OF } (\text{CON-TOP-OF BLOCK}))) \wedge (X_1 \in (\text{GON-TOP-OF}^{-1} (\text{CON-TOP-OF}^{-1} \text{ TABLE})))$

12. PICK-UP-BLOCK-FROM: $(X_2 \in (\text{ON-TOP-OF}^{-1} \text{ BLOCK})) \wedge (X_2 \in (\text{NOT } (\text{CON-TOP-OF}^{-*} \text{ TABLE}))) \wedge (X_2 \in (\text{GON-TOP-OF}^{-*} (\text{ON-TOP-OF}^{-1} \text{ BLOCK}))) \wedge (X_2 \in (\text{GON-TOP-OF}^* (\text{ON-TOP-OF}^{-1} \text{ BLOCK})))$

13. PICK-UP-BLOCK-FROM: $(X_1 \in (\text{GON-TOP-OF}^{-1} (\text{GON-TOP-OF}^{-1} \text{ TABLE})))$

**Boxworld**

1. DRIVE-TRUCK: $(X_2 \in (\text{GBOX-AT-CITY } (\text{BOX-AT-CITY}^{-1} \ X_3))) \wedge (X_3 \in (\text{NOT } (\text{CAN-FLY } (\text{TRUCK-AT-CITY } (\text{NOT PREVIOUS}))))) \wedge (X_3 \in (\text{CAN-DRIVE}^{-1} \text{ PREVIOUS})) \wedge (X_2 \in (\text{NOT } (\text{CAN-FLY } (\text{TRUCK-AT-CITY } (\text{NOT PREVIOUS}))))) \wedge (X_3 \in (\text{NOT } (\text{CAN-FLY } (\text{BOX-AT-CITY BOX})))) \wedge (X_2 \in (\text{CAN-DRIVE } (\text{CAN-DRIVE } (\text{BOX-AT-CITY BOX})))) \wedge (X_3 \in (\text{NOT } (\text{CAN-FLY } (\text{TRUCK-AT-CITY } (\text{BOX-ON-TRUCK } (\text{GBOX-AT-CITY}^{-1} \text{ CITY}))))))$

2. UNLOAD-BOX-FROM-TRUCK-IN-CITY: $(X_1 \in (\text{GBOX-AT-CITY}^{-1} (\text{TRUCK-AT-CITY PREVIOUS}))) \wedge (X_3 \in (\text{GBOX-AT-CITY BOX})) \wedge (X_3 \in (\text{NOT } (\text{BOX-AT-CITY PREVIOUS}))) \wedge (X_1 \in (\text{GBOX-AT-CITY}^{-1} (\text{CAN-DRIVE}^{-1} (\text{CAN-DRIVE}^{-1} (\text{CAN-FLY CITY}))))) \wedge (X_2 \in (\text{BOX-ON-TRUCK } (\text{GBOX-AT-CITY}^{-1} \text{ PREVIOUS})))$

3. DRIVE-TRUCK: $(X_1 \in (\text{BOX-ON-TRUCK } (\text{GBOX-AT-CITY}^{-1} \ X_3))) \wedge (X_2 \in (\text{NOT } (\text{CAN-DRIVE } (\text{TRUCK-AT-CITY } (\text{BOX-ON-TRUCK } (\text{GBOX-AT-CITY}^{-1} \text{ CITY}))))))$

4. DRIVE-TRUCK: ($X_3 \in$ (CAN-DRIVE (BOX-AT-CITY PREVIOUS))) $\wedge$ ($X_2 \in$ (CAN-FLY (CAN-DRIVE$^{-1}$ (BOX-AT-CITY BOX)))) $\wedge$ ($X_3 \in$ (CAN-DRIVE (CAN-FLY (TRUCK-AT-CITY TRUCK)))) $\wedge$ ($X_2 \in$ (NOT (CAN-DRIVE (TRUCK-AT-CITY (BOX-ON-TRUCK (GBOX-AT-CITY$^{-1}$ CITY)))))) $\wedge$ ($X_2 \in$ PREVIOUS) $\wedge$ (CAN-DRIVE (CAN-DRIVE $X_3$))) $\wedge$ ($X_3 \in$ (NOT (TRUCK-AT-CITY (BOX-ON-TRUCK (GBOX-AT-CITY$^{-1}$ CITY))))) $\wedge$ ($X_3 \in$ (NOT (CAN-FLY PREVIOUS))) $\wedge$ ($X_3 \in$ (CAN-DRIVE (NOT (BOX-AT-CITY BOX)))) $\wedge$ ($X_2 \in$ (CAN-DRIVE (CAN-DRIVE$^{-1}$ $X_3$))) $\wedge$ ($X_3 \in$ (CAN-DRIVE (NOT (TRUCK-AT-CITY TRUCK))))

5. LOAD-BOX-ON-TRUCK-IN-CITY: ($X_1 \in$ (GBOX-AT-CITY$^{-1}$ (CAN-DRIVE (TRUCK-AT-CITY TRUCK)))) $\wedge$ ($X_3 \in$ (NOT (PLANE-AT-CITY PREVIOUS))) $\wedge$ ($X_3 \in$ (CAN-DRIVE (CAN-DRIVE$^{-1}$ (CAN-FLY CITY)))) $\wedge$ ($X_3 \in$ (CAN-DRIVE$^{-1}$ (NOT (TRUCK-AT-CITY (NOT PREVIOUS)))))

6. UNLOAD-BOX-FROM-TRUCK-IN-CITY: ($X_3 \in$ (GBOX-AT-CITY (BOX-ON-TRUCK$^{-1}$ TRUCK))) $\wedge$ ($X_3 \in$ (NOT (CAN-FLY (TRUCK-AT-CITY (BOX-ON-TRUCK (GBOX-AT-CITY$^{-1}$ CITY)))))) $\wedge$ ($X_1 \in$ (GBOX-AT-CITY$^{-1}$ CITY))

7. DRIVE-TRUCK: ($X_1 \in$ (BOX-ON-TRUCK (GBOX-AT-CITY$^{-1}$ PREVIOUS))) $\wedge$ ($X_3 \in$ (CAN-DRIVE (GBOX-AT-CITY (GBOX-AT-CITY$^{-1}$ PREVIOUS)))) $\wedge$ ($X_3 \in$ (NOT (PLANE-AT-CITY PLANE))) $\wedge$ ($X_2 \in$ (NOT (CAN-FLY (GBOX-AT-CITY (GBOX-AT-CITY$^{-1}$ PREVIOUS)))))

8. FLY-PLANE: ($X_1 \in$ (BOX-ON-PLANE (GBOX-AT-CITY$^{-1}$ $X_3$)))

9. UNLOAD-BOX-FROM-PLANE-IN-CITY: ($X_1 \in$ (GBOX-AT-CITY$^{-1}$ PREVIOUS))

10. FLY-PLANE: ($X_2 \in$ (NOT (CAN-DRIVE (TRUCK-AT-CITY (BOX-ON-TRUCK (GBOX-AT-CITY$^{-1}$ CITY)))))) $\wedge$ ($X_2 \in$ (GBOX-AT-CITY BOX)) $\wedge$ ($X_3 \in$ (NOT (PLANE-AT-CITY PREVIOUS))) $\wedge$ ($X_1 \in$ (NOT PREVIOUS))

11. LOAD-BOX-ON-PLANE-IN-CITY: ($X_1 \in$ (GBOX-AT-CITY$^{-1}$ (CAN-FLY PREVIOUS))) $\wedge$ ($X_3 \in$ (NOT (TRUCK-AT-CITY (NOT PREVIOUS)))) $\wedge$ ($X_3 \in$ (NOT (CAN-DRIVE (TRUCK-AT-CITY (BOX-ON-TRUCK (GBOX-AT-CITY$^{-1}$ CITY))))))

12. DRIVE-TRUCK: ($X_1 \in$ (BOX-ON-TRUCK (GBOX-AT-CITY$^{-1}$ $X_3$))) $\wedge$ ($X_2 \in$ (NOT (CAN-DRIVE (CAN-FLY PREVIOUS)))) $\wedge$ ($X_2 \in$ (CAN-DRIVE$^{-1}$ (CAN-FLY CITY)))

13. LOAD-BOX-ON-TRUCK-IN-CITY: ($X_1 \in$ (GBOX-AT-CITY$^{-1}$ PREVIOUS))